\documentclass[10pt]{article}
\newif\ifarxiv \arxivtrue

\ifarxiv\usepackage[preprint]{tmlr}\else\usepackage{tmlr}\fi
\usepackage{amsmath,amssymb,booktabs,multirow,graphicx,xcolor}
\usepackage{tikz}
\usetikzlibrary{positioning,arrows.meta,calc,backgrounds,fit}
\usepackage[colorlinks=true,linkcolor=blue!60!black,citecolor=blue!60!black,urlcolor=blue!60!black]{hyperref}

\newcommand{\best}[1]{\textbf{#1}}
\newcommand{\method}{AQLoRA}
\usepackage{xcolor}

\newcommand{\gib}{\ensuremath{\,\mathrm{GiB}}}

\title{\method{}: A Zero-Search Recipe for Fast Quantized LoRA Fine-Tuning}

\ifarxiv
\author{\name Md Romyull Islam \email mislam22@students.kennesaw.edu \\ \addr Kennesaw State University}
\else
\author{\name Anonymous \email anon@example.com \\ \addr Anonymous institution}
\fi

\begin{document}
\maketitle

\begin{abstract}
Quantized fine-tuning (QLoRA) saves memory but not time. It dequantizes every
4-bit weight on the fly, so it trains more slowly than fp16 LoRA. We present
\method{} (Adaptive-Quantization LoRA), a recipe that buys part of that time
back. One CPU pass over the weights fixes both the precision map and the adapter
placement, with no search and no calibration data. The pass ranks layers by NF4
reconstruction error and keeps the top-$K$ in fp16, where $K$ is a fixed
fraction of the linear layers. Those layers skip dequantization,
which is where the speed comes from. A quality setting adapts every layer. A
speed setting adapts only the top blocks, so the backward pass stops early. The rule matches the accuracy of Unsloth's hand-curated dynamic-4bit
selection, in seconds, where search-based allocation needs repeated passes over
calibration data.

We evaluate on Commonsense-170K across six models and four architecture
families, from 1.4B to 14B. The speed setting trains $11.3 \pm 2.3\%$ faster
than well-tuned QLoRA and gives up 0.1 to 1.5 accuracy points. It was faster in
every one of thirteen independent timing sessions, at worst by $7\%$. The quality
setting trains $4.8 \pm 2.0\%$ faster. Its accuracy is level with QLoRA on
every model and within a point of fp16 LoRA. The memory premium depends on
how the architecture packs its layers: 0.06 to 0.8\,GiB on the 1--4B models,
0.5\,GiB at 7B, 1.9\,GiB at 14B. These
error bars are measured between independent sessions, not within one. Three rules follow for timing on shared hardware. Fix the measurement
duration, not the step count. Measure the noise floor from a duplicated arm,
not a nearly identical method. Repeat whole sessions: on one model, three sessions that each passed the floor test spread over $8.2$
points, and the random
seed controls almost none of it.

We validate the recipe with controls and report the two that failed. Choosing
adapter layers by weight density is no better than random. Choosing protected layers by quantization error is not either. We test that on
five models at three budgets, matched on protected memory, and downstream on two
of them. The number of protected layers sets the speed; which layers they are does not
measurably change accuracy.
\end{abstract}

\begin{center}
\small Code and the raw per-run artifacts:
\ifarxiv\url{https://github.com/Romyull-Islam/AQLoRA}\else\url{https://anonymous.4open.science/r/AQLoRA-FC58}\fi
\end{center}

\section{Introduction}
\label{sec:intro}
Low-rank adaptation (LoRA) \citep{hu2022lora} and its quantized form QLoRA
\citep{dettmers2023qlora} made large-language-model fine-tuning cheap on memory.
A frozen 4-bit NF4 backbone with a small adapter fits a billion-parameter model
on one commodity GPU. Memory savings are not speed savings. QLoRA
dequantizes every 4-bit weight block back to half precision on each forward and
backward pass, and that cost makes it slower than fp16 LoRA. Several studies put
NF4 QLoRA at $0.7$ to $0.8$ times the training throughput of plain LoRA
\citep{ko2025loraslower,khaki2025sparselora,qerl2025}. People who quantize to
fit a model at all, whether on an older datacenter card, a laptop, or a phone,
then pay a speed penalty on top of the memory saving.

Part of that penalty can be bought back. An fp16 layer in an
otherwise-quantized model skips dequantization, so every layer kept in fp16
converts memory into speed. On phi-1.5, $0.22$\,GiB of fp16 layers makes
training $6.3\%$ faster than QLoRA (Table~\ref{tab:sessions}). The natural question is which
layers to keep, and the sensitivity literature suggests an answer: NF4 error
concentrates unevenly across layers, so keep the most damaged ones. We adopt that rule and test it. The count of protected layers drives the speed. Which layers are protected does
not measurably change accuracy (\S\ref{sec:ablation-precision}).
The rule survives on different grounds, being deterministic, data-free, and as accurate as a selection industry already ships by hand ($89.0$ against
$89.0$ on Qwen2.5-7B, \S\ref{sec:ablation-precision}). Layer protection itself
is already in use.
Unsloth's dynamic 4-bit models \citep{unsloth2024dynamic} keep chosen layers in
16-bit to restore QLoRA accuracy, and QEFT \citep{lee2024qeft} keeps sensitive
columns in fp16 with custom kernels and reports both quality and speed gains.
What is missing is a stated rule, free of calibration data, that runs on the
stock HuggingFace and bitsandbytes stack with no custom kernels. Also missing
is a comparison against the search-based bit-and-rank allocators
\citep{zhou2025qradaptor,zhou2026autoqra} that define the rest of this
space.

\method{} fills that gap. One CPU pass over the weights takes $5$ to $37$\,s on the models here, with no data
and no forward pass. It produces two maps. A precision map keeps the top-$K$ most NF4-damaged
layers in fp16 within a layer budget. A placement map adapts every layer for
quality, or only the top blocks for speed, so the backward pass stops early. It can be combined with rank-stabilized scaling \citep{kalajdzievski2023rslora},
though that combination is model-dependent (Appendix~\ref{app:rslora}). We apply every generic
speed lever (length-grouped batching, a fused optimizer, the attention backend)
to all baselines alike, so any gain we measure comes from \method{}'s own parts
and not from tuning.

On Commonsense-170K with the 8-task benchmark, across six models and four
architecture families from 1.4B to 14B, \method{}'s quality setting matches
QLoRA everywhere and comes within a point of the fp16 LoRA reference. The same parity holds on GSM8K
math reasoning. Accuracy is a parity result, not a win; the gain is speed.
Sensitivity-based protection and early backward stopping have prior work, which
we cite. Our part is the recipe that composes them, the controls on it, and a
speed result that holds where kernel- and FP8-based methods cannot run. We also checked a
weight-density rule for which layers to adapt, found it no better than random,
and report that. A rank sweep ($r\in\{4,8,16,32\}$) confirms the method ordering is not an
artifact of one adapter size. It also turns up a failure worth flagging:
rsLoRA, the strongest baseline at $r{=}8$, collapses to chance accuracy at
$r{=}32$ under the shared learning rate while its training loss still looks
healthy.

\textbf{Contributions.}
\begin{enumerate}
  \item A \textbf{budgeted protection recipe} for quantized LoRA fine-tuning: one
        calibration-free CPU pass fixes which layers stay in fp16 and which get
        adapters, from a budget on the fraction of layers. Sensitivity-ranked
        protection is established in inference PTQ
        \citep{wu2020integer,dong2020hawqv2,dumitru2024layerwise} and industry
        practice \citep{unsloth2024dynamic}. Ours is the formalization and its evaluation against Unsloth's deployed
        hand-picked selection on accuracy. Unlike search-based allocation
        \citep{zhou2025qradaptor,zhou2026autoqra,guo2024lqlora}, it needs no
        forward pass. The
        ranking inside it is a tie-break we test in item 2, not a contribution we
        claim.
  \item \textbf{Two negative results from our own controls, and the speed
        mechanism that remains}: a weight-density rule for which layers to adapt is no better than random, and
        neither is our NF4-error ranking for which layers to protect. We test the ranking on five models at three budgets against random sets
        matched on protected parameters, and on downstream accuracy on two of
        them. The test also corrected our own control: random draws matched on
        layer count are not matched on memory where layer sizes vary, and that
        artifact had reversed the result on one model. What
        does help is adapting only the top blocks, the known backward-stopping
        speedup \citep{ruckle2021adapterdrop} on a 4-bit base; we measure it and
        do not claim it as new.
  \item A \textbf{systems result on the stock stack}, on unmodified HuggingFace,
        bitsandbytes, and PEFT, timed over thirteen sessions on five models. The
        speed setting trains $11.3 \pm 2.3\%$ faster than well-tuned QLoRA for $0.04$ to $0.8$\,GiB more peak memory,
        giving up $0.1$ to $1.5$ accuracy points. The quality
        setting trains $4.8 \pm 2.0\%$ faster and stays within a point of fp16
        LoRA accuracy. Both are faster than QLoRA in every session.
        This holds on V100-class hardware where FP8 methods
        \citep{choi2025falqon} do not run.
  \item A \textbf{fair-lever evaluation protocol}: every universal efficiency
        lever (length-grouped batching, fused optimizer, dropout, attention
        backend) is applied to \emph{every} baseline, so the measured
        advantage comes only from \method{}'s own mechanisms. We release the code and the raw per-run artifacts, so every number in the
paper can be recomputed without a GPU. The full lever ladder
(Appendix~\ref{app:levers}) makes each lever's contribution
auditable.
\end{enumerate}

\section{Related Work}
\label{sec:related}

\paragraph{Quantized PEFT.} QLoRA \citep{dettmers2023qlora} backpropagates through a frozen NF4 base into
LoRA \citep{hu2022lora}. Successors improve the quantization-aware
initialization (LoftQ \citep{li2024loftq}, ApiQ \citep{liao2024apiq}, CLoQ
\citep{cloq2025}, QERA \citep{zhang2025qera}), mergeability (QA-LoRA
\citep{xu2024qalora}), information retention (IR-QLoRA \citep{qin2024irqlora}),
or quantizer modularity (ModuLoRA \citep{yin2023modulora}, TMLR). None of these targets training \emph{speed}.

\paragraph{Mixed precision and sensitivity-based allocation.} Ranking layers by quantization sensitivity and keeping the worst in higher
precision is long-standing PTQ practice \citep{wu2020integer,dong2020hawqv2,
dumitru2024layerwise}. It is deployed at scale by llama.cpp's k-quant mixes and
by Unsloth's dynamic 4-bit models \citep{unsloth2024dynamic}, which keep chosen
layers in 16-bit for QLoRA fine-tuning based on measured quantization error,
though without a stated selection rule or a systematic study. QEFT
\citep{lee2024qeft} and OWQ \citep{lee2024owq} keep sensitive \emph{columns} in
fp16 with custom layouts and report both quality and fine-tuning speed gains.
Another line assigns per-layer bits and ranks by search: LQ-LoRA uses an ILP
\citep{guo2024lqlora}, LowRA an ILP assigner with learned mappings that reaches
1.15 bits \citep{zhou2025lowra}, QR-Adaptor a gradient-free search
\citep{zhou2025qradaptor}, and AutoQRA an evolutionary and Bayesian search
\citep{zhou2026autoqra}. \method{} differs from these in scope, not idea. It
works at whole-layer granularity on the stock stack rather than QEFT's kernels,
states a budgeted rule where Unsloth ships a hand-picked list, and profiles in $5$ to $37$\,s where the others search.

\paragraph{Adapter placement and selective backpropagation.} That not every layer needs an adapter dates to AdapterDrop
\citep{ruckle2021adapterdrop}, which removed lower-layer adapters and
quantified the resulting backward-pass savings. LoRA-drop
\citep{zhou2024loradrop}, SR-LoRA \citep{srlora2025}, PaFi \citep{liao2023pafi},
and Aletheia \citep{aletheia2026} select layers or parameters by output,
spectral, magnitude, or probe statistics. AdaLoRA \citep{zhang2023adalora}
adapts rank budgets. Stochastic backward reduction
(DropBP \citep{woo2024dropbp}, LISA \citep{pan2024lisa}, LCSB
\citep{lcsb2026}), layer freezing \citep{wang2023egeria,huang2024greentrainer},
and side networks \citep{sung2022lst} all cut backward compute. \method{} takes
depth-restricted placement from this line rather than inventing it, and adds the
measured speed and quality trade-off on a 4-bit base, including how it interacts
with gradient checkpointing.

\paragraph{Fast fine-tuning systems.} Kernel-level systems (Unsloth, Liger,
LoRAFusion \citep{lorafusion2026}) and FP8 methods (FALQON
\citep{choi2025falqon}) set the absolute speed bar on modern datacenter GPUs;
SparseLoRA \citep{khaki2025sparselora} cuts compute via contextual sparsity.
These need custom kernels or Hopper-class hardware. \method{} aims at the other
case, the stock stack on pre-Hopper and edge devices, and its levers still
compose with these systems.

\paragraph{On-device fine-tuning.} Real-phone LLM fine-tuning exists via
memory-efficient BP (MeBP \citep{song2025mebp}), zeroth-order methods (ZeroQAT
\citep{zeroqat2025}), llama.cpp-based mobile LoRA \citep{qvac2025}, and engines
like PockEngine \citep{zhu2023pockengine} and MobileFineTuner
\citep{mobilefinetuner2025}; EDGE-LLM \citep{yu2024edgellm} adapts compression
in simulation. All fix base precision by a uniform setting or a fixed heuristic
mix; none selects per-layer precision from a measurement of the model being
tuned.

\section{Method}
\label{sec:method}

\begin{figure}[t]
  \centering
  \includegraphics[width=\textwidth]{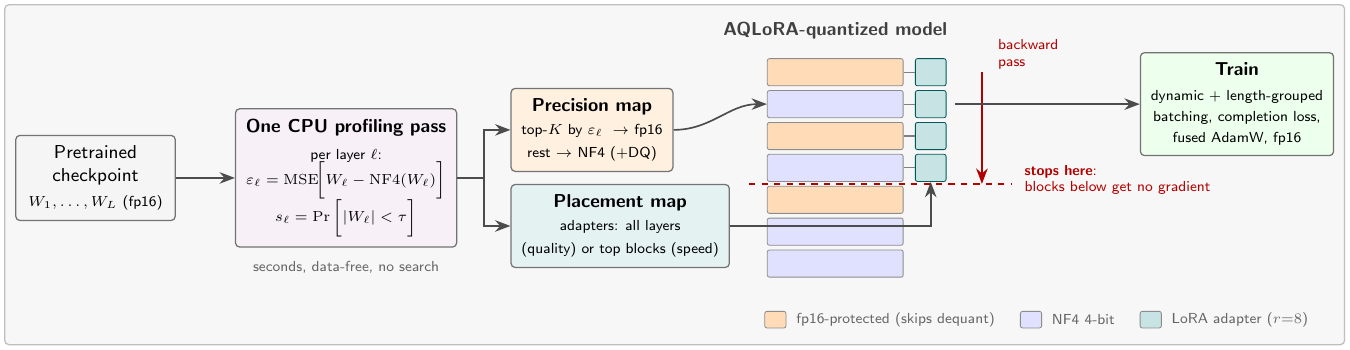}
  \caption{The \method{} pipeline.}
  \label{fig:arch}
\end{figure}

\subsection{Profiling pass}
For each linear layer $\ell$ with weights $W_\ell$, flattened into blocks
$b$ of 64 elements (matching bitsandbytes),
\begin{equation}
  \varepsilon_\ell \;=\; \frac{1}{|W_\ell|}\sum_{b}\big\lVert b -
  \mathrm{absmax\text{-}NF4}^{-1}\!\big(\mathrm{absmax\text{-}NF4}(b)\big)\big\rVert_2^2,
  \qquad
  s_\ell \;=\; \Pr\big[\,|W_{\ell,ij}| < \tau\,\big],
\end{equation}
with $\tau{=}0.01$ and the true 16-level NF4 quantile grid. The pass runs on CPU over the fp16 checkpoint, is data-free, and requires no
forward pass. On 40 CPU threads with the checkpoint in the page cache, it takes $5$\,s on the
1B models, $12$ to $13$\,s at 3 to 4B, $25$\,s at 7B and $37$\,s at 14B; reading
the checkpoint from disk adds up to $21$\,s (released as
\texttt{profile\_timing.json}). This is negligible beside the multi-hour fine-tuning run. The calibration- and search-based allocation of
QR-Adaptor and AutoQRA \citep{zhou2025qradaptor,zhou2026autoqra} instead
evaluates each candidate configuration with forward passes on calibration data.

\subsection{Precision allocation (zero-search)}
The errors $\varepsilon_\ell$ are far from uniform. On phi-1.5 they span $11.8\times$ across the 144 linear layers, and the top fifth
hold $29\%$ of the total (Figure~\ref{fig:sensitivity}, Appendix~\ref{app:acc}), so a small fp16 budget can cover
the worst offenders. Given a protection budget $\rho \in [0,1]$
(\texttt{sensitive-frac}), the $K{=}\lceil\rho L\rceil$ layers with largest
$\varepsilon_\ell$ are loaded in fp16; all others in NF4 with double
quantization; \texttt{lm\_head} stays fp16. Note that $\rho$ counts layers, not
parameters: the same $\rho$ buys $4.3\%$ of the weights on Llama-3.2-1B and
$16.3\%$ on Phi-3-mini (Table~\ref{tab:arch}), so it is a budget on the rule's input
and not directly on memory. The rule is a fixed top-$K$ cut with no calibration data, no search, and no
ILP. We call this \emph{zero-search} in a specific sense: the configuration is
fixed by one deterministic pass over the weights, with no evaluation of
candidate configurations. It is not zero-cost. The pass itself costs the $5$ to $37$\,s measured above, and the budget $\rho$ remains a hyperparameter we
set once and do not tune per model. It helps speed, since fp16
layers skip the per-step dequantization that is the main reason QLoRA runs
slower than fp16 LoRA \citep{ko2025loraslower,khaki2025sparselora}.

We state one caveat here rather than leaving it to the ablations, because it
concerns the rule itself. Ranking by $\varepsilon_\ell$ is the natural reading of
the sensitivity literature, and it is what we implemented, but when we test it
against parameter-matched random selections it is not measurably ahead
(\S\ref{sec:ablation-precision}). The count of fp16 layers is what buys the speed, and on the evidence we have the identity of those layers is not what buys
the quality. We keep the rule because it is deterministic, data-free, costs $5$ to $37$\,s, and
matches the accuracy of a deployed hand-curated selection, not because we can
show it beats picking $K$ layers arbitrarily.

\subsection{Adapter placement}
An adapter budget $\alpha$ (\texttt{adapter-frac}) sets how many of the $L$
layers get an adapter. Adapting fewer layers is faster but costs accuracy ($0.7$ points at $\alpha{=}.7$ on phi-1.5, \S\ref{sec:ablation-placement}), so $\alpha$ trades speed against accuracy. We considered choosing the $\lceil\alpha
L\rceil$ layers by weight density ($s_\ell$), but this does no better than a
random subset (\S\ref{sec:ablation-placement}), so we do not use it: the quality
setting adapts every layer. For speed we instead use \textbf{depth} placement,
which puts adapters on the top blocks only. No trainable parameter then sits
below the cut block, so autograd stops the backward pass there, and with
non-reentrant gradient checkpointing the lower blocks also skip recomputation.
This is the AdapterDrop mechanism \citep{ruckle2021adapterdrop}, and unlike a
random subset it makes the saving concrete rather than incidental.

\subsection{Composition and levers}
\method{} can be combined with rank-stabilized scaling
\citep{kalajdzievski2023rslora} ($+$rs rows); Appendix~\ref{app:rslora} shows
the combination is model-dependent. All experiments also use a set of shared levers, applied to every
method: SDPA attention, LoRA dropout 0, length-grouped dynamic batching, and a
fused AdamW. The lever ladder (Table~\ref{tab:ladder}) measures each one.

We report two settings of \method{}, both with $\rho{=}.2$ and no rsLoRA scaling
(the rsLoRA composition is optional and model-dependent, \S\ref{sec:rslora}).
\textbf{\method{}-q} is the quality setting: adapters on every layer
($\alpha{=}1.0$). \textbf{\method{}-s} is the speed setting: depth placement with
$\alpha{=}.85$ and early backward stopping. These are two settings of one budget, not separate methods.

\section{Experimental setup}
\label{sec:setup}
\textbf{Protocol.} Commonsense-170K fine-tuning with the Alpaca template and
completion-only loss, evaluated by greedy generation on the eight held-out
test sets (BoolQ, PIQA, SIQA, HellaSwag, WinoGrande, ARC-e, ARC-c, OBQA)
following \citet{hu2023llmadapters,liu2024dora}. Secondary: WikiText-2
perplexity of the quantized base (Table~\ref{tab:precision}).
\textbf{Models.} phi-1.5 (1.4B), Llama-3.2-1B, Llama-3.2-3B,
Phi-3-mini-4k-instruct (3.8B), Qwen2.5-7B, and Qwen2.5-14B: four architecture
families across a $10\times$ size range, chosen so the memory regime shifts from
``quantization optional'' to ``quantization required'' within the study.

The two mid-scale models are there for architectural variation, not for another
point on the size axis. The rule operates on the list of linear layers, and that
list is not the same shape in every family
(Table~\ref{tab:arch}, Appendix~\ref{app:acc}). Grouped-query attention shrinks \texttt{k\_proj} and
\texttt{v\_proj} well below \texttt{q\_proj}, and Phi-3 fuses its projections
into \texttt{qkv\_proj} and \texttt{gate\_up\_proj}, leaving four kinds of linear
layer where Llama has seven. A budget of $\rho{=}.20$ therefore buys
different things: $4.3\%$ of the weights on Llama-3.2-1B against $16.3\%$ on
Phi-3-mini, a factor of $3.8$ for the same nominal setting, and a factor of
fourteen in the memory it costs. We report this because it is a defect in how we
specified the budget. A fraction of \emph{layers} does not transfer across
architectures; a fraction of \emph{parameters} would. Comparisons at equal $\rho$
across families are therefore comparisons at unequal memory, and we give Phi-3 a
memory-matched budget alongside the nominal one for that reason.

\textbf{Baselines.} Quantized: QLoRA, rsLoRA, QDoRA, LoftQ; fp16 references:
LoRA and DoRA. LoftQ runs on three of the six models and fails on Llama-3.2-3B
and Qwen2.5-7B for an environment reason, not a property of the method; we do
not read its absence in our favour (Appendix~\ref{app:fidelity}). PiSSA
cannot initialize on a quantized base by design \citep{meng2024pissa}, and LoRA+
and AdaLoRA vary the learning-rate ratio or rank budget, a setting independent of
precision, so we leave all three out of this study. Search-based joint allocation
\citep{zhou2025qradaptor,zhou2026autoqra} has no public code; we compare
allocation cost analytically, and compare against Unsloth's released selection
directly in \S\ref{sec:ablation-precision}.
\textbf{Hyperparameters.} $r{=}8$, LoRA scaling factor $16$ (that is, $2r$),
lr $3{\times}10^{-4}$
cosine, 1 epoch, batch $6\times 8$ accumulation, 600 tokens, fp16.
\textbf{Fidelity.} Each baseline's settings were checked against its official
repository; deviations and per-method notes are Appendix~\ref{app:fidelity}.
\textbf{Hardware.} One V100-SXM2-32GB per run, a pre-Hopper datacenter GPU. All
results in this paper are on this card. We target smaller commodity and edge
GPUs, but do not run on them here; \S\ref{sec:edge} gives a feasibility analysis
with bandwidth-scaled projections, not measurements.
\textbf{Timing protocol.} Accuracy and timing are measured separately. Accuracy
does not depend on load, so accuracy runs share the node with other users; their
wall-clock is not a timing measurement, and no table reports it. For timing, each
model's methods run in one session on one GPU, \emph{interleaved} so that any
drift in machine state affects every method equally, with five or seven repeats
per method and 200 measured steps after warmup (60 in the two earliest kept sessions). In twelve of the thirteen sessions, each run starts only once the
timing GPU is empty; the earliest phi-1.5 session instead held the node's other
GPUs idle for its duration. The node is shared and not reserved. NVML confirms
that the timing GPU held its maximum clock ($1530$\,MHz) with no thermal or power
throttling, including while other users' jobs ran on other GPUs of the same
chassis. Interference can only slow a run, so the samples have a ceiling at the
uncontended speed and a slow tail. We therefore estimate speed as the mean of the
fastest half of the repeats: three or four runs rather than one lucky maximum, without the interference-slowed
tail.

\textbf{Noise floor and exclusion.} Each session measures its own noise floor.
Nine sessions include a second, identically configured QLoRA arm, and the floor is
the gap between the two QLoRA arms. The four earliest kept sessions predate that arm
and use a proxy instead: rsLoRA and LoftQ differ from QLoRA only in a scaling
constant and in initialization, so their gap to QLoRA is also a floor. A session
is kept when its floor is under $3\%$; at or above that, the floor is the size of
the effect being measured. Five sessions fail, with floors of $3.5$ to $8.7\%$,
and are excluded; all five are in the released artifacts. Every model is timed
in two or three kept sessions, thirteen in all. Appendix~\ref{app:sens} repeats
the analysis under three other estimators, with the excluded sessions included,
and with the duplicate-arm sessions alone. The lever ladder
(Appendix~\ref{app:leverfig}) is a single sequential pass and does not follow
this protocol.
\textbf{Allocator.} All runs set PyTorch's \texttt{expandable\_segments}
allocator, applied identically to every method, so it lowers absolute peak memory
without affecting any comparison between methods.
\textbf{Efficiency metrics.} Steps/s and real (non-pad) tokens/s (the timing
protocol), plus padding waste, peak and base-model GiB, trainable parameters,
and profiling seconds. We treat efficiency as a primary result, next to accuracy.
\textbf{Artifacts.} Every run writes a \texttt{report.json} with its per-task scores, peak memory,
wall-clock, and the exact plan used. All of them are released with the code, so
each table can be recomputed from the artifacts alone.

\section{Main results}
\label{sec:main}

\begin{table}[t]
\centering\small
\caption{Commonsense-170K 8-task accuracy (\%) on phi-1.5, no rsLoRA scaling.}
\label{tab:main-phi15}
\setlength{\tabcolsep}{3.2pt}
\resizebox{\textwidth}{!}{%
\begin{tabular}{lcccccccc|c|cc}
\toprule
Method & BoolQ & PIQA & SIQA & HellaS & WinoG & ARC-e & ARC-c & OBQA & Avg & Peak\gib & Params \\
\midrule
LoRA fp16 (ref) & \best{65.0} & 76.5 & 76.5 & 83.2 & 78.0 & 80.5 & 67.1 & 71.6 & 74.8 & 5.18 & 7.1M \\
DoRA fp16 (ref) & 64.6 & 76.5 & \best{76.9} & 83.1 & \best{78.4} & 81.2 & \best{67.6} & 71.8 & 75.0 & 5.19 & 7.5M \\
\midrule
QLoRA$^{\dagger}$ & 64.5 & 76.6 & 76.5 & 83.2 & 76.9 & 81.3 & 66.6 & 71.9 & 74.7 & \best{3.83} & 7.1M \\
rsLoRA$^{\dagger}$ & 64.7 & \best{77.3} & 76.3 & \best{83.5} & 77.4 & \best{81.5} & 66.5 & \best{74.0} & \best{75.2} & \best{3.83} & 7.1M \\
LoftQ$^{\ddagger\S}$ & 64.6 & 76.3 & 76.3 & 83.4 & 77.6 & \best{81.5} & 67.0 & 72.1 & 74.8 & \best{3.83} & 7.1M \\
QDoRA           & \best{65.0} & 76.2 & 76.0 & 82.9 & 76.8 & 81.4 & 67.2 & 70.6 & 74.5 & 3.84 & 7.5M \\
\midrule
\method{}-q$^{\dagger}$ & 64.7 & 76.7 & 76.2 & 83.2 & 77.1 & 81.2 & 66.7 & 72.5 & 74.8 & 4.05 & 7.1M \\
\method{}-s     & 64.7 & 74.4 & 75.4 & 82.0 & 75.4 & 79.0 & 65.3 & 69.6 & 73.2 & 4.03 & 6.0M \\
\bottomrule
\end{tabular}}\\[2pt]
{\scriptsize Avg = mean of the eight task columns; bold = best per column on
unrounded values. $^{\dagger}$Mean of six seeds, $^{\S}$of three
(Table~\ref{tab:seeds}); unmarked rows are single runs. $^{\ddagger}$PEFT
one-shot NF4 initialization. Speed is in Tables~\ref{tab:cleantime} and~\ref{tab:sessions}.
}
\end{table}

\begin{figure}[t]
  \centering
  \begin{minipage}{0.54\textwidth}\centering
    \includegraphics[width=\textwidth]{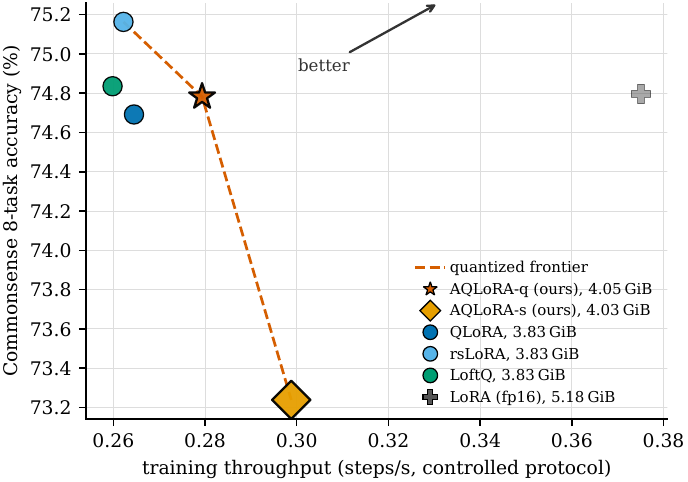}
  \end{minipage}\hfill
  \begin{minipage}{0.44\textwidth}\centering
    \includegraphics[width=\textwidth]{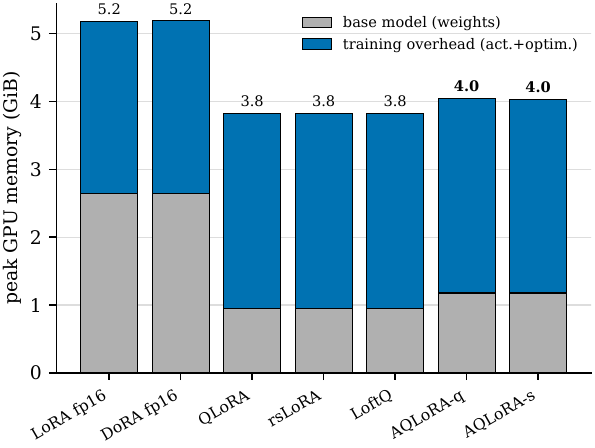}
  \end{minipage}
  \caption{Left: accuracy against training throughput on phi-1.5; accuracy as in
  Table~\ref{tab:main-phi15}, throughput from Table~\ref{tab:cleantime}. Right:
  peak memory.}
  \label{fig:pareto}
\end{figure}

On phi-1.5 every method except \method{}-s sits in a tight band (74.5 to 75.2 average). Over
repeated runs \method{}-q ($74.78$), LoftQ ($74.84$) and QLoRA ($74.69$) sit
within $0.15$ of one another and of the fp16 LoRA reference ($74.80$); only
rsLoRA separates, at $75.16$ (Table~\ref{tab:seeds}). The gaps among the first four
are smaller than the seed spread of any one of them, so the result is parity, not a win. The central rows are repeated because a single seed can overstate a margin, as
the next case shows. Llama-3.2-1B
makes the point concretely (Table~\ref{tab:main-llama}). At one seed \method{}-q
led QLoRA by $1.1$ points; over three seeds the margin is $0.3$ ($72.2$ against
$71.9$), because QLoRA's first seed was its worst. \method{}-q
is level with LoftQ ($72.3$, one run) there. Across both models \method{}-q matches fp16 LoRA within seed noise and
is at or above QLoRA, at a fraction of fp16's memory (base $1.18$ against $2.64$\,GiB on phi-1.5) and higher throughput than QLoRA (Table~\ref{tab:sessions}). \method{}-s (depth $\alpha{=}.85$) trades accuracy for speed. Against QLoRA's
multi-seed means it is $1.5$ points lower on phi-1.5 ($73.2$), $1.0$ on
Llama-3.2-3B ($82.25$ against $83.25$), $0.4$ on Phi-3-mini, $0.3$ on
Llama-3.2-1B ($71.6$ against $71.9$), and $0.1$ on Qwen2.5-7B. At seed 42 alone
it leads QLoRA on Llama-3.2-1B ($71.6$ against $71.1$), but that is QLoRA's
worst seed. The speed setting costs $0.1$ to $1.5$ points. QDoRA reaches QLoRA accuracy under these settings; under its own recipe it gains
$0.89$ points and passes \method{}-q by $0.26$ (Appendix~\ref{app:parity}). The closest competitor is LoftQ. The two are tied on accuracy: $74.84$ for LoftQ over three seeds against
$74.78$ over six, inside the spread of either. LoftQ uses less memory, $3.83$
against $4.05$\,GiB of peak, since the fp16-protected layers cost $0.22$\,GiB. \method{}-q is faster, by $4.8 \pm 2.0\%$ over QLoRA-class methods
across sessions (Table~\ref{tab:sessions}). So \method{}-q is behind or level on two of the three axes a practitioner
weighs, and its case rests on step time alone. A reader who does not care about training time should prefer
LoftQ here. Both
sit well under fp16 LoRA's $5.18$\,GiB (Figure~\ref{fig:pareto}, right). The
accuracy is not the selling point; the speed at parity is (\S\ref{sec:levers}).

Table~\ref{tab:main-3b} (Appendix~\ref{app:acc}) gives Llama-3.2-3B and Phi-3-mini on the 8-task benchmark, three seeds per
cell. Neither changes the accuracy picture. On Llama-3.2-3B the quantized methods
sit between $81.7$ and $83.5$ against an fp16 reference of $84.0$, and
\method{}-q's $+0.26$ over QLoRA carries $p{=}.32$. On Phi-3-mini the three main
4-bit methods span $0.07$ points, QLoRA at $87.0$, LoftQ at $87.0$ and
\method{}-q at $87.0$, with $p{=}.73$ against QLoRA. We read the second as the useful one. On a current architecture at 3.8B, the
choice among quantized LoRA methods does not move this benchmark. Differences of this size are smaller than the spread between repeated runs of
one configuration (\S\ref{sec:seeds}).

The memory column is where the two models separate, and it is the reason we added
them. \method{}-q costs $+0.27$\,GiB over QLoRA on Llama-3.2-3B and $+0.82$ on
Phi-3-mini for the same $\rho$, which is the fused-projection effect of
Table~\ref{tab:arch} showing up as measured peak rather than as an estimate.
Peak memory is not comparable \emph{across} the two: Llama-3.2 carries a
128K-token vocabulary against Phi-3's 32K, which by itself accounts for most of
the $3.8$\,GiB difference in non-weight memory, and has nothing to do with
precision allocation. Only the premium within a model is a property of the method.

\subsection{Scale: where quantization stops being a choice}
\label{sec:scale}
At 1.4B, fp16 LoRA fits everywhere and there is no reason to quantize; the
methods above compete in a band best read as parity. Scale changes
that. Table~\ref{tab:scale} runs the core set on Qwen2.5-7B and reports memory
at 14B, same protocol throughout. At 7B, fp16 LoRA peaks at 19.1\,GiB, which
already crowds a 24\,GB consumer card, while \method{}-q matches its accuracy
(89.0 for both) at 10.6\,GiB. At 14B, fp16 LoRA is out of reach on commodity
hardware: its weights alone are 27.5\,GiB, past a 24\,GB card, while the
quantized methods load the same base in 9 to 11\,GiB. In this regime
quantization is not an optimization but a requirement, and the question becomes
which quantized method to use. \method{} does not undercut QLoRA on memory; the fp16-protected layers add 0.5\,GiB at 7B and 1.9\,GiB at 14B over uniform
4-bit. At 7B that premium buys a measured speedup (Table~\ref{tab:cleantime}).
At 14B, \method{}-s reaches slightly higher accuracy than QLoRA (91.3 against
91.1); 14B was not timed under the controlled protocol, so we make no speed claim
there. At 7B, \method{}-q is above QLoRA on all three seeds
(\S\ref{sec:seeds}). rsLoRA underperforms at the shared learning rate (87.1,
\S\ref{sec:rslora}). 

\begin{table}[t]
\centering\small
\caption{Scale study: Qwen2.5-7B accuracy and the 14B memory picture.}
\label{tab:scale}
\begin{tabular}{lc|c}
\toprule
Qwen2.5-7B & Avg & Peak\gib \\
\midrule
LoRA fp16 (ref) & \best{89.0} & 19.06 \\
\method{}-q     & 89.0 & 10.55 \\
\method{}-s     & 88.7 & 10.51 \\
QLoRA           & 88.8 & \best{10.04} \\
rsLoRA          & 87.1 & \best{10.04} \\
\bottomrule
\end{tabular}\\[5pt]
\begin{minipage}{0.92\textwidth}\small
Rows are single runs at seed 42; \S\ref{sec:seeds} gives three seeds for
\method{}-q ($89.08$) and QLoRA ($88.80$). LoftQ fails on Qwen2.5-7B for an
environment reason
(Appendix~\ref{app:fidelity}); DoRA and QDoRA were run on the two small models
only. \emph{Qwen2.5-14B memory (32\,GB V100).} fp16 LoRA needs $27.5$\,GiB for the
weights alone, past a 24\,GB card and near a 32\,GB V100's limit; that run was
stopped early (43-h ETA) and has no accuracy entry. QLoRA reaches $91.1$ at $12.4$\,GiB,
\method{}-s $91.3$ at $14.2$\,GiB, and \method{}-q $91.1$ at $14.3$\,GiB. The two
\method{} rows compute the protected layers in fp32, with storage unchanged
(\S\ref{sec:limitations}).
\end{minipage}
\end{table}

\paragraph{A second task family.}
We also check that the parity result is not specific to commonsense data. We
fine-tune Qwen2.5-7B on a 20K subset of MetaMathQA and score exact-match
accuracy on the GSM8K test set (1{,}319 problems), the MetaMathQA-to-GSM8K protocol of PiSSA
\citep{meng2024pissa}. The same picture holds (Table~\ref{tab:gsm8k}, Appendix~\ref{app:acc}). The three methods land within $0.5$ points of one another, inside the
$\pm 1$-point sampling noise of this test set. The quantized runs peak at $10.05$ and $10.55$\,GiB against fp16's $19.06$. The
parity claim carries across task families; here too the difference between
methods is cost, not accuracy.

\section{Where the speed comes from}
\label{sec:levers}

Table~\ref{tab:sessions} reports speed over the thirteen kept sessions.
\method{}-s averages $11.3\%$ faster than QLoRA with a between-session standard
deviation of $2.3$, and it is faster in every session, by $7.0$ to $15.2\%$; the
smallest margin exceeds every floor. \method{}-q averages $4.8 \pm 2.0\%$ and is
also faster in every session, by $2.1$ to $8.1\%$. Neither result depends on the
estimator or the exclusions (Appendix~\ref{app:sens}). rsLoRA and LoftQ share
QLoRA's step cost to within the floor (Table~\ref{tab:cleantime}), so the same
margins hold against them. fp16 LoRA is faster still wherever it fits, by $16.6$
to $41.8\%$ over QLoRA in Table~\ref{tab:cleantime}.

A floor measured inside one session does not predict the spread between
sessions. On phi-1.5, Llama-3.2-1B and Qwen2.5-7B, repeated sessions of
\method{}-s land within $0.1$, $0.7$ and $0.2$ points of one another. On
Llama-3.2-3B the two later sessions sit $5.5$ and $6.4$ points above the first.
On Phi-3-mini three kept sessions span $8.2$ points, from $7.0$ to $15.2$, while
one of them reports a within-session floor of $1.3\%$. Excluded sessions differ
more: on phi-1.5 they put \method{}-s at $16.7$ and $6.2\%$, against $13.0$ in a
kept session. A speed number from a single session, here or elsewhere, carries
an uncertainty that its own error bar does not show. Table~\ref{tab:cleantime}
is kept beside Table~\ref{tab:sessions} to show the difference.

Per-model means of \method{}-s range from $10.5\%$ (Qwen2.5-7B) to $12.9\%$
(phi-1.5). Those two models repeat tightly and differ from each other (Welch
$p{<}.001$). The other three spread by up to $8$ points across sessions, and a
one-way test across all five models does not separate them ($p{=}.89$). The gain
may depend on the model; with two or three sessions each, these data cannot say
how.

For timing on shared hardware this reduces to three rules, each tied to a failure
documented in Appendix~\ref{app:forensics}. Fix the measurement duration, not
the step count, or faster models get noisier numbers. Measure the floor from two
runs of the same configuration; a nearly identical method reads too narrow.
Repeat the whole session before quoting a margin; a clean floor is necessary, not
sufficient. The seed does not control this variance (\S\ref{sec:seeds}).

The lever ladder (Table~\ref{tab:ladder}) adds one change at a time on phi-1.5 in
a single sequential pass, so its margins are single-session numbers.
Length-grouped batching is the largest shared lever, cutting padding waste from
$39.8\%$ to $3.7\%$. Truncating the backward pass at the lowest adapted block
adds $5.6\%$ at $\alpha{=}.7$ ($0.287$ to $0.303$ steps/s). With gradient checkpointing off, depth placement at $\alpha{=}.7$ runs $42\%$
faster than QLoRA at $23\%$ less
memory, because stopping the backward pass early also drops the lower-block
activations.

On accuracy against throughput alone, fp16 LoRA and rsLoRA form the frontier
(Figure~\ref{fig:pareto}). Within the 4-bit memory class, both settings of ours
are on the frontier and QLoRA and LoftQ are not, though the accuracy gaps are
inside seed noise. The figure labels each point with its peak memory, since the
frontier is only meaningful within a class.

\begin{table}[tbp]
\centering\small
\caption{Steps/s from one session per model under the controlled protocol.}
\label{tab:cleantime}
\begin{tabular}{lcccccc|c}
\toprule
& QLoRA & rsLoRA & LoftQ & \method{}-q & \method{}-s & LoRA fp16 & floor \\
\midrule
phi-1.5$^{a}$    & 0.265 & 0.262 & 0.260 & 0.279 & \best{0.299} & 0.375 & $1.77\%$ \\
\quad vs.\ QLoRA & ref & $-0.9\%$ & $-1.8\%$ & $+5.6\%$ & \best{$+13.0\%$} & $+41.8\%$ & \\
\midrule
Llama-3.2-1B$^{c}$ & 0.372 & 0.366 & 0.371 & 0.381 & \best{0.414} & 0.492 & $1.78\%$ \\
\quad vs.\ QLoRA & ref & $-1.8\%$ & $-0.5\%$ & $+2.4\%$ & \best{$+11.1\%$} & $+32.1\%$ & \\
\midrule
Llama-3.2-3B$^{c}$ & 0.213 & 0.210 & n/a$^{b}$ & 0.219 & \best{0.229} & 0.265 & $\mathbf{2.24\%}^{d}$ \\
\quad vs.\ QLoRA & ref & $-1.8\%$ & n/a$^{b}$ & $+2.5\%$ & \best{$+7.2\%$} & $+24.3\%$ & \\
\midrule
Phi-3-mini$^{c}$ & 0.254 & 0.257 & 0.257 & 0.259 & \best{0.271} & 0.296 & $1.33\%$ \\
\quad vs.\ QLoRA & ref & $+1.3\%$ & $+1.2\%$ & $+2.1\%$ & \best{$+7.0\%$} & $+16.6\%$ & \\
\midrule
Qwen2.5-7B$^{a}$ & 0.194 & 0.194 & n/a$^{b}$ & 0.201 & \best{0.215} & 0.237 & $0.10\%$ \\
\quad vs.\ QLoRA & ref & $-0.1\%$ & n/a$^{b}$ & $+3.6\%$ & \best{$+10.5\%$} & $+21.9\%$ & \\
\bottomrule
\end{tabular}\\[2pt]
{\scriptsize Seven interleaved repeats per method; estimate = mean of the fastest
half. $^{a}$60 measured steps; $^{c}$200 (\S\ref{sec:setup}). $^{b}$LoftQ unavailable
(Appendix~\ref{app:fidelity}); floor rests on rsLoRA alone.
$^{d}$Duplicate-arm floor; the other four are proxy floors and read narrow
(\S\ref{sec:setup}). Percentages are comparable within a row only. The phi-1.5
\method{} arms carry rsLoRA scaling, a constant factor with no compute cost;
rsLoRA itself, which differs from QLoRA only by that factor, lands $0.9\%$ away.
}
\end{table}

\begin{table}[tbp]
\centering\small
\caption{Speed over independent sessions, percent faster than QLoRA.}
\label{tab:sessions}
\begin{tabular}{lccc}
\toprule
Model & clean sessions & \method{}-q & \method{}-s \\
\midrule
phi-1.5        & 2 & $+6.3 \pm 0.9$ & $+12.9 \pm 0.0$ \\
Llama-3.2-1B   & 2 & $+5.1 \pm 3.9$ & $+11.4 \pm 0.5$ \\
Llama-3.2-3B   & 3 & $+4.4 \pm 1.7$ & $+11.1 \pm 3.5$ \\
Phi-3-mini     & 3 & $+4.8 \pm 3.0$ & $+11.1 \pm 4.1$ \\
Qwen2.5-7B     & 3 & $+4.0 \pm 0.4$ & $+10.5 \pm 0.1$ \\
\midrule
pooled         & 13 & \best{$+4.8 \pm 2.0$} & \best{$+11.3 \pm 2.3$} \\
\bottomrule
\end{tabular}\\[2pt]
{\scriptsize Sessions with a floor of $3\%$ or more are excluded (five; \S\ref{sec:setup}).}
\end{table}

\section{Ablations}
\label{sec:ablations}
\subsection{Which layers to adapt}
\label{sec:ablation-placement}
Picking the layers to adapt by weight density does not help. We test it at two
budgets ($\alpha{\in}\{0.5,0.7\}$) against a weight-norm ranking (the Act-LoRA
baseline) and a random subset (Table~\ref{tab:e3}). The table has two columns that
differ in seed and in rsLoRA scaling, which the seed-42 runs used; comparisons
hold within a column. Random is ahead of density in all four cells, by $1.0$ and
$1.1$ points at $\alpha{=}.5$ and by $0.2$ and $0.5$ at $\alpha{=}.7$, and density
is last of the three in every cell. The density score gives no benefit and we do
not use it: the quality setting adapts every layer.

What matters more is the budget. In the seed-42 column, dropping from every layer to $\alpha{=}.7$ costs $0.7$ points ($75.56$ to $74.84$). In the lever ladder,
cutting to $\alpha{=}.7$ with depth placement but no backward truncation raises
throughput from $0.260$ to $0.287$ steps/s (Table~\ref{tab:ladder}), so most of
that gain comes from the adapter count. Restricting the adapters to the top
blocks (depth) costs more accuracy at the same budget ($71.2$ at $\alpha{=}.7$);
in exchange, truncating the backward pass adds $5.6\%$ throughput. At
$\alpha{=}.85$ the depth rule scores $74.6$ in that column. Without rsLoRA scaling,
which is the shipped speed setting, it scores $73.2$ (Table~\ref{tab:main-phi15}).

\IfFileExists{fig_placement.pdf}{%
}{}

\subsection{Precision budget}
\label{sec:ablation-precision}
Base perplexity improves with the protection budget under both NF4 and the
harsher fp4 (Appendix~\ref{app:pplbudget} prices this). That perplexity gain does not carry to the downstream task. The budget $\rho{=}.20$ was chosen on this evidence. Table~\ref{tab:rho-sweep}
then sweeps it end to end and reports 8-task accuracy, memory, and step time. Accuracy shows no trend: the single-seed budgets span $0.6$ points, less than the
$0.76$ that seeds 42 to 44 span at $\rho{=}.20$ alone ($74.35$ to $75.11$). Base
perplexity keeps falling where accuracy does not: the residual tier of
Appendix~\ref{app:residual} at $\rho{=}.40$ reaches $18.83$, against $18.94$ for
fp16 at $\rho{=}.20$, and scores $74.96$. Base perplexity, which we used to choose
the budget, is a poor guide to it.

Since accuracy is flat here, the budget should be set by cost. At $\rho{=}.05$ the
premium over QLoRA falls to $0.06$\,GiB and in one session the step-time advantage is $4.1\%$ against a $2.8\%$ noise
floor. The flat result does not hold on
Phi-3-mini, where $\rho{=}.05$ scores lower, so it is not a default
(Appendix~\ref{app:budget}).

\paragraph{Does the ranking select better than chance?} On phi-1.5, accuracy does not respond to how many layers are protected. The next
question is whether it responds to \emph{which} ones. The speed side cannot answer it, because any $K$ fp16 layers skip the same
dequantization. We hold the protected memory fixed and vary only the selection:
top-$K$ by NF4 error, the $K$ least damaged layers, and ten random draws matched
to top-$K$. Base perplexity needs no training and is deterministic given the
layer set, so these numbers are exact.

Matching the draws by layer count, as our first control did, is not enough.
Where layer sizes vary, the most damaged layers are smaller than average ($5.5$M
against $8.4$M parameters on phi-1.5), so a random draw of the same count carries
more parameters. On
Llama-3.2-1B the draws held up to $11\%$ more memory than top-$K$ and won on
perplexity for that reason alone. We therefore match on parameter total and let
the count float. That holds the protected memory equal to within $0.1\%$ in every
setting reported here, and it is the control we recommend for any budgeted
allocation rule.

Under that control the ranking has no measurable advantage
(Appendix~\ref{app:nullfull} lists
every setting). Across five models and three budgets, top-$K$ beats the median
matched draw in $11$ of $15$ settings, and $50$ of the $150$ draws beat top-$K$
where chance predicts $75$. A sign test does not separate that from chance
($p{=}.12$), and the median gap is $0.01$ perplexity. The direction is mildly
favourable and the size is negligible.

Downstream the selections are indistinguishable on both models we tested. On
phi-1.5 at seed 42, top-$K$ and matched random score $74.88$ and $74.63$, against
a seed spread near $0.25$ (Table~\ref{tab:seeds}); bottom-$K$, which protects
$79\%$ more parameters, scores $74.89$. On Phi-3-mini, where the perplexity gap
is widest, top-$K$ and matched random reach $87.04$ and $87.03$ over three seeds.
Phi-3-mini is also the model on which no quantized method separates
(\S\ref{sec:main}), so phi-1.5 is the more discriminating test, at one seed.

What protection buys, speed and lower base perplexity, therefore comes from
protecting some layers, not from protecting the right ones. NF4 reconstruction error is not doing the work we
attributed to it. This does not touch the speed result, which depends only on how
many layers avoid dequantization, nor the accuracy parity. It does remove the mechanism we offered: that protecting the most damaged layers
is what preserves quality. A reader should treat the
rule as a cheap and reproducible way to fix a layer count, not as a sensitivity
criterion shown to work. A parameter-matched random selection does as well; the ranking costs $5$ to $37$\,s of CPU time and gives a fixed, documented choice, which is
the only reason it remains the default. The test is at whole-layer granularity, so it says
nothing about the column-granularity selection of QEFT and OWQ.

A cheaper protection tier, a second NF4 stage over each protected layer's own
residual, also works and is also not worth using; Appendix~\ref{app:residual}
reports it as a negative result.

We also compare our rule against the only deployed alternative we know of:
Unsloth's released dynamic-4bit selection for Qwen2.5-7B, a hand-curated list covering 34 of the 196 linear layers. Substituting their list for ours in an otherwise identical run gives 89.0 on
the 8-task average, matching our 89.0. The two selections are not the same
layers: their list names 10 modules, ours 30, and only 6 are shared. Two
different selections at different budgets landing on the same accuracy is the
ranking control (Appendix~\ref{app:nullfull}) seen from the other side. The practical point survives it: the budgeted rule needs no per-model
curation to match the hand-tuned selection.

The rank sweep, the rsLoRA composition, and the edge estimate are in
Appendices~\ref{app:rank}, \ref{app:rslora}, and \ref{app:edge}; the method
ordering holds at every rank, and the composition does not reach significance.

\subsection{Why the two rules fail}
\label{sec:why}
The density and weight-norm rules pick almost the same layers: 68 of 72 at
$\alpha{=}.5$ and 99 of 100 at $\alpha{=}.7$. The density score is therefore close
to a weight-magnitude ranking, and where the two sets differ by one layer they
score within $0.1$ to $0.3$ points of each other. The sets differ from a random
draw in two ways. They sit higher in the network: their mean adapted block is
$12.5$ to $13.4$ of 0 to 23, against $10.8$ to $11.7$ for random draws
(Table~\ref{tab:e3}). They also skew by module type: at $\alpha{=}.5$ the density
rule adapts all 24 query projections and one of 24 value projections, where a random draw takes $9$ to $12$ of each.

To separate the two, we drew random layers with the density rule's count in each
third of the network. The depth profile then matches, while the layers do not
(36 to 40 of 72 shared). In both columns of Table~\ref{tab:e3} this
depth-matched control lands $0.25$ points below unconstrained random and $0.8$ to
$0.9$ above density: $74.3$ against $74.6$ and $73.6$ with rsLoRA scaling, $73.3$
against $73.5$ and $72.4$ without. Depth accounts for $23$ to $25\%$ of the density rule's loss. The rest comes from which layers it picks; module type is
the likely cause, since the depth-matched sets spread across types as random does,
but we did not test it separately.

The depth rule is the extreme case of moving adapters up. At $\alpha{=}.7$ its mean
block is $15.2$ and it scores $3.7$ points below density at the same budget. In
exchange, no layer below its lowest adapted block needs a gradient, so the
backward pass stops there. The density rule moves adapters up without emptying the
lowest blocks, so it pays an accuracy cost and gets no backward-pass cut.

The error ranking fails for two reasons that the corrected control separates.
First, NF4 error falls with layer size ($r{=}-0.19$): the $29$ most damaged
layers on phi-1.5 average $5.5$M parameters against $8.4$M overall. A control
matched on count therefore hands the random draws more memory, which is the
artifact of \S\ref{sec:ablation-precision}. Second, the base-perplexity advantage
that survives a fair control is at most $0.05$ against the median matched draw,
and it does not reach accuracy. That is consistent with the premise of LoftQ
\citep{li2024loftq}, that low-rank adapters compensate quantization error: a
difference this small between two protected sets is within what the adapter
absorbs. Protection pays in speed, by removing dequantization work, and lowers base perplexity as the budget grows (Appendix~\ref{app:pplbudget}). Which layers carry the
fp16 does not measurably matter once the model is fine-tuned.

\subsection{Seeds}
\label{sec:seeds}
On Qwen2.5-7B we reran the central pair on three seeds: \method{}-q reaches $89.08 \pm 0.19$ against QLoRA's $88.80 \pm 0.11$, above it on every seed with
no overlap (worst \method{}-q seed 88.97, best QLoRA seed 88.91). A gap of $0.28$ points is why we call the accuracy parity; its sign is the same
on every seed.

On phi-1.5 the same pair ties (Table~\ref{tab:seeds}): \method{}-q averages $74.78 \pm 0.25$ against QLoRA's $74.69 \pm 0.22$ over the six runs of Table~\ref{tab:seeds}, so the seed-42 gap ($74.88$ against
$74.74$) does not survive repetition. At three seeds
\method{}-q also looked far less reproducible ($F{=}71$, two-sided $p{=}.03$); a fourth seed
per arm brought that to $F{=}1.9$ ($p{=}.60$), and six seeds to $F{=}1.2$. A variance ratio from three runs per arm can flip on one draw.

These error bars do not measure seed variation. Three runs of \method{}-q at the \emph{same} seed spread as widely as six runs
at different seeds (sd $0.24$ against $0.25$). Our training does not enable deterministic kernels, so the error bars measure
run-to-run nondeterminism, and fixing the seed does not make results reproducible.

\section{Limitations}
\label{sec:limitations}
The largest limitation is the selection rule. It ranks layers by NF4
reconstruction error. Tested against parameter-matched random selections on five
models at three budgets, it is not measurably ahead, and the advantage it does
show on base perplexity does not reach downstream accuracy
(\S\ref{sec:ablation-precision}). Protecting layers buys speed and lowers base perplexity; protecting the ones our
criterion picks is not measurably better than protecting an arbitrary
set of the same size. The speed result is unaffected, since it depends on how
many layers avoid dequantization rather than which. The mechanism offered for the recipe is not supported by the control. A reader should take \method{} as a reproducible way, costing $5$ to $37$\,s, to fix how
many layers are protected, not as a demonstration that sensitivity ranking works at
whole-layer granularity. The test covers five models at three budgets each, ten draws per setting. That
is enough to say the criterion earns nothing measurable at whole-layer
granularity. It does not rule out a gain at finer granularity, which we did not
test.

At 14B some of the layers our rule protects overflow in fp16 backward on
pre-bf16 GPUs. This silently collapses the AMP
loss scaler while the training loss still looks plausible. Uniform QLoRA is
immune, and our mitigation computes the protected layers in fp32 at unchanged
storage.
On bf16 hardware the issue does not arise.
We do not expect \method{} to win at matched memory. Paying for the
fp16-protected layers by shrinking the batch or sequence trades the memory saving
back, so the gain is speed, not accuracy per byte. The speed setting gives up $0.1$ to $1.5$ points of accuracy against QLoRA
(\S\ref{sec:main}), so its speed is not free. Our speedups are on
the stock stack and pre-Hopper or edge GPUs; kernel-level and FP8 methods are
faster where their hardware applies,
and while \method{}'s levers should compose with them, we do not measure that
here. The phone numbers are projections from prior measured data, not on-device
runs. Our headline tables fix one epoch, and the rank sweep
(\S\ref{sec:rank}) covers $r\in\{4,8,16,32\}$ on phi-1.5 and $r\in\{8,16,32\}$ on
Llama-3.2-1B, not the larger models.

\section{Conclusion}
\method{} sets per-layer precision and adapter placement for quantized LoRA from
one CPU pass over the weights, with no search. Across six models and four
architecture families its accuracy matches QLoRA's and trails fp16 LoRA by less
than a point, and across thirteen timing sessions its speed setting trains $11.3 \pm 2.3\%$ faster than well-tuned QLoRA, for $0.04$ to $0.8$\,GiB more peak memory on the
timed models. The
result is speed at parity, not higher accuracy. No single part is new:
protection by sensitivity, selective placement, and early backward stopping all
have prior work, which we cite. What we add is a recipe that composes them from one data-free pass, and the
controls that show which part of it does the work. It matches the accuracy of a
hand-curated industrial selection where search-based allocators need repeated
calibration passes. It also yields three measurement findings. How many layers are protected sets the speed; which ones does not measurably
change accuracy. A noise floor measured inside one session understates the spread between
sessions: on Phi-3-mini, a $1.3\%$ floor against an $8.2$-point span. The seed controls almost none of the run-to-run variance. The speed result
holds on the stock stack, on hardware where kernel- and FP8-based methods do not
run.  A measured phone-GPU system remains
future work.

\subsubsection*{Broader Impact Statement}
\method{} lowers the time and hardware needed to fine-tune a language model on
older and smaller GPUs; the main benefit is access for groups without current
datacenter hardware, and faster training means less energy per run. The same
access cuts both ways, since cheaper fine-tuning is cheaper for harmful
adaptation too. We see no risk specific to this work beyond those that apply to
efficient fine-tuning in general.

\bibliography{refs}
\bibliographystyle{tmlr}

\appendix
\section{On the shared levers and the ranking}
\label{app:parity}
Table~\ref{tab:main-phi15} gives every method the same shared settings (fused
AdamW, dropout 0, length-grouped batching), which do not match each method's own
published recipe. To test whether that choice drives the result, we reran the
four contested methods on phi-1.5 under their native settings instead: paged
8-bit optimizer, dropout $0.05$, no length grouping, everything else unchanged
(Table~\ref{tab:parity}).

Every method scores higher under its native recipe, so the shared levers cost
each of them accuracy: $0.27$ to $0.89$ points. For the three methods that share a compute path
(QLoRA, LoftQ, \method{}-q) the shift is small and nearly uniform, $+0.27$ to
$+0.40$ points, and their order is unchanged. QDoRA is the exception: it gains
$+0.89$, more than twice as much, and moves from last under the shared levers to
second under its own recipe, passing \method{}-q by $0.26$ points. So the shared
levers are not neutral for every method, and a reader should not read
Table~\ref{tab:main-phi15} as a recipe-independent ranking. \method{}-q stays at or above QLoRA in both settings, which is the comparison it rests
on. QDoRA was not timed under the controlled protocol, so we make no speed
comparison with it.

\begin{table}[tbp]
\centering\small
\caption{Native recipe against the shared levers, phi-1.5 (8-task avg, \%).}
\label{tab:parity}
\begin{tabular}{lccc}
\toprule
Method & native & shared levers & change \\
\midrule
LoftQ       & \best{75.53} & \best{75.13} & $+0.40$ \\
QDoRA       & 75.41 & 74.52 & $+0.89$ \\
\method{}-q & 75.15 & 74.88 & $+0.27$ \\
QLoRA       & 75.08 & 74.74 & $+0.34$ \\
\bottomrule
\end{tabular}\\[2pt]
{\scriptsize Both columns are seed 42, so the pairing holds
within each row; the shared-lever column is a single run, not the multi-seed
mean of Table~\ref{tab:main-phi15}.}
\end{table}

\section{Additional accuracy tables and figures}\label{app:acc}
\begin{table}[tbp]
\centering\small
\caption{Llama-3.2-3B and Phi-3-mini, 8-task accuracy (\%), mean of three seeds.}
\label{tab:main-3b}
\setlength{\tabcolsep}{3.2pt}
\resizebox{\textwidth}{!}{%
\begin{tabular}{lcccccccc|c|cc}
\toprule
Method & BoolQ & PIQA & SIQA & HellaS & WinoG & ARC-e & ARC-c & OBQA & Avg &
Peak\gib & Params \\
\midrule
\multicolumn{12}{l}{\emph{Llama-3.2-3B}} \\
LoRA fp16 (ref) & \best{71.6} & \best{86.4} & \best{80.8} & \best{95.1} & \best{86.4} & \best{89.5} & \best{77.7} & \best{84.1} & \best{84.0} & 11.96 & 12.2M \\
QLoRA & 70.8 & 85.8 & 80.5 & 94.7 & 84.9 & 89.2 & 77.4 & 82.6 & 83.3 & \best{8.07} & 12.2M \\
rsLoRA & 69.2 & 83.8 & 79.8 & 93.3 & 83.3 & 87.0 & 74.6 & 82.9 & 81.7 & \best{8.07} & 12.2M \\
\method{}-q & 70.6 & 86.1 & 80.7 & 94.8 & 85.6 & 89.1 & 77.4 & 83.9 & 83.5 & 8.33 & 12.2M \\
\method{}-s & 69.1 & 85.0 & 80.3 & 94.4 & 85.2 & 87.6 & 75.9 & 80.5 & 82.2 & 8.30 & 10.3M \\
\midrule
\multicolumn{12}{l}{\emph{Phi-3-mini}} \\
LoRA fp16 (ref) & 71.3 & \best{88.4} & 81.8 & \best{94.9} & \best{86.3} & \best{96.2} & \best{88.9} & \best{90.9} & \best{87.3} & 9.31 & 12.6M \\
QLoRA & 70.9 & 88.1 & \best{81.9} & 94.6 & 85.8 & 95.8 & 88.5 & 90.1 & 87.0 & \best{4.30} & 12.6M \\
LoftQ & 70.8 & 88.3 & 81.8 & 94.7 & 86.0 & 96.1 & 88.4 & 89.5 & 87.0 & 4.34 & 12.6M \\
rsLoRA & \best{71.7} & 88.1 & 81.7 & 94.6 & 85.9 & 95.8 & 87.3 & 89.8 & 86.9 & \best{4.30} & 12.6M \\
\method{}-q & 71.0 & 88.3 & 81.5 & 94.7 & 85.7 & 96.1 & 88.2 & 90.8 & 87.0 & 5.12 & 12.6M \\
\method{}-s & 70.3 & 87.5 & 81.1 & 94.5 & 85.6 & 95.8 & 88.0 & 89.7 & 86.6 & 5.09 & 10.6M \\
\bottomrule
\end{tabular}}\\[2pt]
{\scriptsize LoftQ is absent on Llama-3.2-3B for the environment reason given
above; DoRA and QDoRA were run on phi-1.5 and Llama-3.2-1B only. Peak memory is comparable within a model, not between the two: the
vocabulary sizes differ by $4\times$.}
\end{table}
\begin{table}[tbp]
\centering\small
\caption{GSM8K accuracy (\%) after MetaMathQA-20K fine-tuning, Qwen2.5-7B.
Three methods were run on this task, one
run each.}
\label{tab:gsm8k}
\begin{tabular}{lc|c}
\toprule
Method & GSM8K & Peak\gib \\
\midrule
QLoRA           & \best{84.9} & \best{10.05} \\
LoRA fp16 (ref) & 84.8 & 19.06 \\
\method{}-q     & 84.4 & 10.55 \\
\bottomrule
\end{tabular}
\end{table}
\begin{figure}[t]
  \centering
  \includegraphics[width=0.62\textwidth]{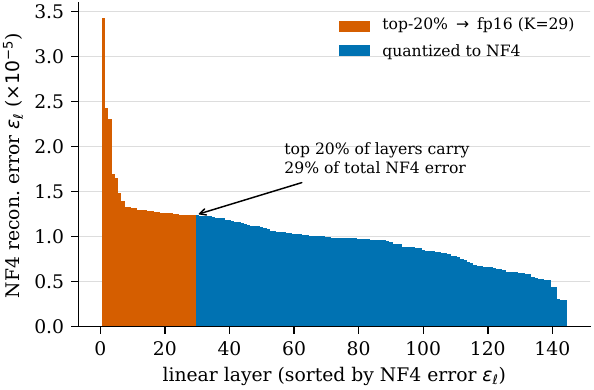}
  \caption{Sorted per-layer NF4 error on phi-1.5. Orange: layers kept in fp16.}
  \label{fig:sensitivity}
\end{figure}
\begin{table}[tbp]
\centering\small
\caption{What the allocation rule sees in each family.}
\label{tab:arch}
\begin{tabular}{lccccc}
\toprule
Model & linear layers & layer kinds & size CV & \% params at $\rho{=}.20$ & premium\gib \\
\midrule
phi-1.5           & 144 & 6 & 0.71 & 13.2\% & 0.220 \\
Llama-3.2-1B      & 112 & 7 & 0.82 & \best{4.3\%} & \best{0.058} \\
Llama-3.2-3B      & 196 & 7 & 0.67 & 6.8\% & 0.265 \\
Phi-3-mini (3.8B) & 128 & 4 & 0.52 & 16.3\% & 0.817 \\
\bottomrule
\end{tabular}\\[2pt]
{\scriptsize Size CV is the coefficient of variation of layer parameter counts;
premium is fp16 storage for the top-$K$ set over all-NF4.}
\end{table}
\begin{table}[t]
\centering\small
\caption{Commonsense-170K 8-task accuracy (\%) on Llama-3.2-1B, no rsLoRA
scaling.}
\label{tab:main-llama}
\setlength{\tabcolsep}{3.2pt}
\resizebox{\textwidth}{!}{%
\begin{tabular}{lcccccccc|c|cc}
\toprule
Method & BoolQ & PIQA & SIQA & HellaS & WinoG & ARC-e & ARC-c & OBQA & Avg & Peak\gib & Params \\
\midrule
LoRA fp16 (ref) & 63.8 & 78.2 & 74.6 & \best{87.6} & \best{71.8} & \best{77.3} & 59.4 & \best{72.4} & \best{73.1} & 7.81 & 5.6M \\
DoRA fp16 (ref) & \best{64.9} & \best{78.5} & 73.9 & 86.8 & 71.0 & 76.8 & 59.6 & 70.2 & 72.7 & 7.82 & 6.0M \\
\midrule
QLoRA$^{\dagger}$ & 63.2 & 77.9 & 74.2 & 85.9 & 70.8 & 75.6 & 58.9 & 68.7 & 71.9 & \best{6.47} & 5.6M \\
LoftQ           & 63.5 & 77.6 & 74.6 & 86.2 & 71.6 & 75.6 & 59.1 & 69.8 & 72.3 & 6.58 & 5.6M \\
QDoRA           & 63.2 & 78.4 & 74.1 & 85.9 & 70.8 & 75.3 & \best{59.9} & 69.8 & 72.2 & 6.68 & 6.0M \\
\midrule
\method{}-q$^{\dagger}$ & 63.8 & 76.9 & \best{74.7} & 86.4 & 70.8 & 75.8 & 59.0 & 70.1 & 72.2 & 6.53 & 5.6M \\
\method{}-s     & 63.7 & 77.0 & 73.7 & 85.8 & 70.2 & 75.1 & 59.3 & 68.4 & 71.6 & 6.51 & 4.8M \\
\bottomrule
\end{tabular}}\\[2pt]
{\scriptsize Conventions follow Table~\ref{tab:main-phi15}. $^{\dagger}$Mean of
three seeds; unmarked rows are single runs. rsLoRA is omitted: on this model
every rsLoRA-scaled run fell to 65 to 67 at the shared learning rate (rsLoRA
itself $66.6$; Appendix~\ref{app:rslora}).}
\end{table}

\section{Rank sweep}\label{app:rank}

\label{sec:rank}
The results so far fix $r{=}8$. Table~\ref{tab:rank} repeats the quantized set at
$r\in\{4,8,16,32\}$ on phi-1.5, same protocol throughout. Two things hold across
the sweep and one baseline breaks.

\method{}-q is at or above QLoRA at all four ranks, though never by much: $+0.26$
at $r{=}16$, $+0.03$ at $r{=}8$ over three seeds, and $+0.01$ at $r{=}32$. The
parity result is therefore not an artifact of one rank, and neither is its size.
Its memory premium over QLoRA is constant at $0.22$\,GiB, which is what the rule
predicts: the fp16-protected set is fixed, while the adapters grow with rank, so
the premium falls from $5.7\%$ of peak at $r{=}8$ to $5.3\%$ at $r{=}32$. LoftQ
matches \method{}-q within noise at all four ranks, differing by at most $0.1$
points and in both directions; we read that as a tie, not a win for either, and
note that LoftQ reaches it at $0.22$\,GiB less peak memory.
Rank itself matters more than the choice among these methods: the working methods gain $1.7$ to $1.9$ points from $r{=}4$ to $r{=}32$, a larger effect than any gap
between them at a fixed rank.

That sweep is phi-1.5 only, so we repeated $r\in\{16,32\}$ on Llama-3.2-1B
(Table~\ref{tab:rank-llama}, one seed per cell). Parity reproduces at $r{=}16$:
\method{}-q, QLoRA and LoftQ land within $0.5$ points, with \method{}-q $0.33$
below QLoRA, inside QLoRA's seed spread on this model ($0.71$, Table~\ref{tab:main-llama}). Two things differ from phi-1.5. Rank does not help here,
since every method is lower at $r{=}16$ than at $r{=}8$, the opposite of the
phi-1.5 trend. And at $r{=}32$ QLoRA and \method{}-q both fall to chance while
LoftQ degrades but survives at $63.2$. The collapse is inherited from the
baseline rather than caused by the protection scheme, and it has the same origin
as the rsLoRA case: the runs finish with normal eval loss at a rate tuned for
$r{=}8$. We therefore scope the rank conclusions to phi-1.5 and treat $r{=}32$ on
Llama-3.2-1B as unusable at the shared learning rate. LoftQ's robustness there is
real and we do not explain it.

The exception is rsLoRA, which is the strongest method at $r{=}4$ ($74.78$) and
$r{=}8$ ($75.16$ over six seeds), then degrades at $r{=}16$ ($73.86$) and
collapses to chance at $r{=}32$
($37.47$; HellaSwag $25.7$, ARC-c $23.8$ against a $25$ floor). The cause is the scaling, not the method. rsLoRA divides the LoRA scaling
factor by $\sqrt{r}$ instead of by $r$, so with our factor of $2r$ the update
is $2.8\times$ larger at $r{=}8$, $4\times$ at $r{=}16$, and $5.7\times$ at
$r{=}32$. At our shared learning rate the largest of these
overshoots. This is the same failure we report
on Llama-3.2-1B (\S\ref{sec:rslora}), here reached by raising the rank instead of
changing the model. Two practical notes follow. A method that wins at one rank
under a shared learning rate can lose badly at another, so single-rank
comparisons are fragile. And the collapse is invisible in the training signal:
the $r{=}32$ run ends at eval loss $0.108$, close to its $r{=}16$ value of
$0.103$, while its task accuracy is at chance. Held-out loss is not a safe proxy
for downstream accuracy here.

\begin{table}[tbp]
\centering\small
\caption{Rank sweep on phi-1.5 (8-task avg, \%).}
\label{tab:rank}
\begin{tabular}{lcccc}
\toprule
Method & $r{=}4$ & $r{=}8$ & $r{=}16$ & $r{=}32$ \\
\midrule
\method{}-q & 74.2 & 74.8 & \best{75.5} & 75.9 \\
QLoRA       & 74.1 & 74.8 & 75.2 & 75.9 \\
rsLoRA      & \best{74.8} & \best{75.3} & 73.9 & 37.5$^{\ast}$ \\
LoftQ       & 74.1 & 74.8 & 75.5 & \best{76.0} \\
\bottomrule
\end{tabular}\\[2pt]
{\scriptsize $^{\ast}$Chance-level collapse from rsLoRA scaling at the shared
learning rate, not a crash: the run finished with a normal eval loss ($0.108$).
The $r{=}8$ column is the mean of seeds 42 to 44 (Table~\ref{tab:seeds}; Table~\ref{tab:main-phi15}
extends three of these rows to six seeds); the other ranks are seed 42.
}
\end{table}

\begin{table}[tbp]
\centering\small
\caption{Rank on Llama-3.2-1B, 8-task average (\%), one seed except $r{=}8$,
which is the multi-seed mean of Table~\ref{tab:main-llama} where available.}
\label{tab:rank-llama}
\begin{tabular}{lccc}
\toprule
Method & $r{=}8$ & $r{=}16$ & $r{=}32$ \\
\midrule
LoftQ       & \best{72.27} & \best{70.06} & \best{63.20} \\
QLoRA       & 71.91 & 69.88 & 37.23 \\
\method{}-q & 72.20 & 69.55 & 38.57 \\
rsLoRA      & 66.58 & 37.40 & 36.89 \\
\bottomrule
\end{tabular}
\end{table}

\section{The rsLoRA composition}\label{app:rslora}
\label{sec:rslora}

\paragraph{Budget under the composition.} The recommendation does not extend to the rsLoRA composition, and the evidence
does not show it fails there either. Repeating the two cheap budgets with $+$rs gives $75.46$ at
$\rho{=}.05$ and $75.13$ at $\rho{=}.10$, against $75.50$ over five seeds at
$\rho{=}.20$. Against the three-seed standard deviation of that arm ($0.06$) both read as
clearly worse. At five seeds the standard deviation is $0.35$ and both sit within
one of the $\rho{=}.20$ mean, so the comparison separates nothing. One seed per
budget against an arm this variable is not evidence. The composition stays at
$\rho{=}.20$ for want of evidence, not because the cheaper budgets were shown to
fail.

\method{} can be composed with rsLoRA scaling \citep{kalajdzievski2023rslora}.
The effect is model-dependent, and on six seeds per arm it does not reach
significance. On phi-1.5 \method{}$+$rs averages $75.49 \pm 0.31$ against the
rsLoRA baseline's $75.16 \pm 0.37$, a margin of $0.32$ with $p{=}.14$ and
overlapping ranges (Table~\ref{tab:seeds}).

How that number moved is the more useful result. At three seeds the margin was
$0.29$ with $p{=}.005$ and no overlap between runs. At six seeds it is $0.32$
with $p{=}.14$. The effect size barely changed. Three seeds put both standard
deviations near $0.06$; six put them near $0.35$. The significance was in an
underestimate of the spread, not in the effect, and nothing in the three-seed
version signalled it.

We therefore do not claim a win here. The margin is positive on the seeds we
have, it is of the same size as the spread of the arms producing it, and its
scope was already narrow: the same scaling collapses on Llama-3.2-1B (below) and
at $r{=}32$ on phi-1.5 (\S\ref{sec:rank}). A practitioner should read $+$rs as a
composition worth trying with a tuned learning rate, not as a property of
\method{}. On Llama-3.2-1B it hurts badly. rsLoRA scaling divides the LoRA scaling factor by $\sqrt{r}$ instead of by
$r$, which enlarges the update $2.8\times$ at $r{=}8$. At our learning
rate of $3\times10^{-4}$ that overshoots on Llama: every rsLoRA-scaled run
(\method{}$+$rs, rsLoRA, \method{}-s$+$rs) fell to
65 to 67, while the same methods without the scaling trained normally (the
$71$--$73$ range of Table~\ref{tab:main-llama}). We did not tune the learning rate per method, so these runs show the failure,
not whether a smaller rate would avoid it. The rank sweep
(\S\ref{sec:rank}) reaches the same failure from the other direction: on phi-1.5,
where $+$rs helps at $r{=}8$, raising the rank to $32$ enlarges the same factor to
$5.7\times$ and drives the rsLoRA baseline to chance. We therefore do
not scale by default and treat $+$rs as an optional add-on that needs its own
learning rate per model and per rank.

\begin{table}[tbp]
\centering\small
\caption{Seeds 42 to 47 on phi-1.5, 8-task average (\%).}
\label{tab:seeds}
\setlength{\tabcolsep}{4pt}
\begin{tabular}{lcccccc|c}
\toprule
Method & 42 & 43 & 44 & 45 & 46 & 47 & mean $\pm$ std \\
\midrule
QLoRA          & 74.74 & 74.71 & 74.80 & 74.30 & 74.62 & 74.97 & 74.69 $\pm$ 0.22 \\
LoftQ$^{\dagger}$ & 75.13 & 74.65 & 74.73 & -- & -- & -- & 74.84 $\pm$ 0.26 \\
\method{}-q    & 74.88 & 75.11 & 74.35 & 74.78 & 74.75 & 74.82 & 74.78 $\pm$ 0.25 \\
rsLoRA         & 75.40 & 75.29 & 75.27 & 74.44 & 75.11 & 75.46 & 75.16 $\pm$ 0.37 \\
\method{}$+$rs & 75.56 & 75.61 & 75.67 & 75.75 & 74.89 & 75.42 & \best{75.49 $\pm$ 0.31} \\
\bottomrule
\end{tabular}\\[2pt]
{\scriptsize $^{\dagger}$LoftQ was run on seeds 42 to 44 only.}
\end{table}

\section{Edge feasibility}\label{app:edge}
\label{sec:edge}
This section is a feasibility estimate, not a result: the only measurements in it
are on the V100, and every device figure is a bandwidth-scaled projection. It bounds what the recipe would need on hardware not available to us, and it
should not be read as more than that.
A phone-sized run (batch 1, 256 tokens, rank 4) fine-tunes phi-1.5 in
1.44\,GiB on a V100 at 232 tokens/s. That is 24\% above QLoRA's throughput, at
1.44 against QLoRA's smaller 1.24\,GiB, and well within the 8 to 24\,GB of a
2026 flagship phone (Table~\ref{tab:edge}). At batch 1 the work is bound by memory
bandwidth, not compute: the run reaches $1.29$ TFLOPS, $1\%$ of the card's
$125$ TFLOPS fp16 peak, because it reads the weights once per token. So the right
thing to scale by, going to a phone, is DRAM bandwidth. The V100 has $900$\,GB/s
and the two phones in Table~\ref{tab:edge} have $80$ and $85$\,GB/s, a factor of
$10.6$ to $11.3$, which puts a phone at $1.2$ to $1.3$\,h for a 1000-example epoch. QVAC reports 1\,h 40\,min per epoch
for a larger 1.7B model on an Adreno 830 \citep{qvac2025}, an order-of-magnitude
check on the projection rather than a match (their epoch is over a different
dataset, so the two are not directly comparable). These are upper bounds; better mobile
kernels will lower them. The on-device system, with per-tensor GGUF precision
matching our precision map and a llama.cpp or Metal training path
\citep{qvac2025,song2025mebp}, is future work.

\begin{table}[tbp]
\centering\small
\caption{Edge profile (phi-1.5, batch 1, 256 tokens, rank 4). Device rows are projections.}
\label{tab:edge}
\begin{tabular}{lcccc}
\toprule
Device & DRAM BW & proj.\ tok/s & peak GiB & 1k-example epoch \\
\midrule
V100 (\textbf{measured}), AQLoRA-edge & 900 GB/s & 232 & 1.44 & 6.8 min \\
V100 (\textbf{measured}), QLoRA-edge & 900 GB/s & 187 & 1.24 & 8.4 min \\
\midrule
\multicolumn{5}{l}{\emph{bandwidth-bound projection for AQLoRA-edge (upper bound; on-device kernel maturity will lower it):}} \\
Jetson AGX Orin (LPDDR5) & 205 GB/s & $\sim$53 & 1.44 & $\sim$0.5 h \\
iPhone 16 Pro (LPDDR5X) & 85 GB/s & $\sim$22 & 1.44 & $\sim$1.2 h \\
OnePlus 15 / SD 8 Elite (LPDDR5X) & 80 GB/s & $\sim$21 & 1.44 & $\sim$1.3 h \\
\bottomrule
\end{tabular}

\end{table}

\section{Timing forensics: two ways a sweep fails}\label{app:forensics}
One detail invalidated four sweeps, and it is easy to repeat. The first protocol
timed every model over the same number of steps. A fixed step count is
not a fixed measurement: Llama-3.2-1B is the fastest model here at $0.37$ steps/s,
so 60 measured steps took $164$\,s, while the same 60 steps on Qwen2.5-7B took
$297$\,s. The same absolute jitter is then a much larger fraction of the shorter
measurement, and the noise floor followed the window length rather than anything
about the machine: $0.1\%$ at $297$\,s, $1.8\%$ at $217$\,s, and $3.1\%$ at
$164$\,s. Four Llama-3.2-1B sweeps failed the threshold and were discarded. The cause was
the short window, not contention: Holding the \emph{duration}
fixed instead, at 200 measured steps for a $524$\,s window, brought the floor to
$1.78\%$ on the first attempt, and the resulting numbers agree with the discarded
ones to within a third of a point. We therefore set the step count per model to
target a comparable measurement window, and we recommend that anyone timing
models of different speeds do the same.

The window is not the only thing that widens a floor. A later sweep on Llama-3.2-3B ran a $1006$\,s
window on a card that read $20$\,MiB before all $42$ runs, and still came out at
$5.15\%$. Its per-repeat floors were $20.7$, $11.3$, $2.3$, $1.7$, $2.8$, $8.1$
and $0.3$ percent, and every method slowed together partway through, which is
what interference looks like and what a method difference does not. We discarded it, and a
later sweep on a quieter node reached $2.24\%$ and is the one reported in
Table~\ref{tab:cleantime}. The two failure modes are separate:
too short a window makes the noise unmeasurable, and a busy machine makes it
large. Only the first is under the experimenter's control, which is why the floor
has to be measured every time rather than assumed from an earlier session.

\section{Precision budget and base perplexity}\label{app:pplbudget}
\begin{table}[tbp]
\centering\small
\caption{Precision budget and base WikiText-2 perplexity, phi-1.5.}
\label{tab:precision}
\begin{tabular}{lcccc}
\toprule
Quantizer & fp16 & $\rho{=}0$ (uniform) & $\rho{=}.10$ & $\rho{=}.20$ \\
\midrule
NF4 (+DQ)   & 18.52 & 18.97 & 18.97 & \best{18.94} \\
fp4 (no DQ) & 18.52 & 19.43 & 19.38 & \best{19.37} \\
\midrule
fp16 layers & all & 0 & 14 & 29 \\
base (GiB)  & 2.64 & 0.95 & 1.05 & 1.18 \\
\bottomrule
\end{tabular}
\end{table}
A larger protection budget $\rho$ lowers quantization damage on the base model.
Table~\ref{tab:precision} gives the WikiText-2 perplexity of the quantized model
with no adapters, so it measures the quantization alone, as $\rho$ grows, under
NF4 and under the harsher fp4 without double quantization. Keeping $20\%$ of the
layers in fp16 recovers a small, consistent part of the gap to fp16, and the
effect is larger under fp4, where the budget matters most. Section~\ref{sec:ablation-precision} shows that this recovery depends on the
budget, not on which layers are selected. The table also
prices the budget: $\rho{=}.20$ moves 29 of 144 layers to fp16 and grows the base
from $0.95$ to $1.18$\,GiB, the $0.22$\,GiB premium that shows up again as peak memory in every phi-1.5
table.

\section{A residual protection tier}\label{app:residual}
Protection can also be made cheaper instead of smaller. Storing a protected layer
as a second NF4 stage over its own residual, rather than in fp16, removes 110
times the protected-layer reconstruction error for $8.25$ bits per parameter
against fp16's $16$; halving the block size, which also adds scales, removes only
$1.1$ times. It works downstream and is still not worth using: $74.97$ at
$3.91$\,GiB peak, where plain fp16 at $\rho{=}.05$ scores the same at $3.89$, and
it gives up $3.6\%$ step time because a protected layer then dequantizes twice
instead of not at all. Given a memory budget, protecting fewer layers exactly
beats protecting more layers approximately.

\section{Baseline fidelity notes}\label{app:fidelity}
LoRA and DoRA. LoftQ runs on phi-1.5, Llama-3.2-1B and Phi-3-mini, and fails on
Llama-3.2-3B and Qwen2.5-7B, where PEFT's initialisation cannot resolve the
checkpoint on disk under our offline cache and raises before any weights are
touched. We report it as an environment failure rather than a property of the
method, because it does not follow model size and does not follow checkpoint
layout: Phi-3-mini is sharded exactly as the two failing models are and
initialises normally, so a single-file checkpoint requirement does not explain
it either. A fix would need network access to re-fetch the affected snapshots,
which the cluster does not have. LoftQ
is the closest competitor to \method{} in this study and is level with it
wherever both run, so its absence at two scales is a gap in our evidence and we
do not read it in our own favour. 

\textbf{Fidelity notes} (each baseline's settings were checked against its
official repository). All methods share the protocol above, which differs from some
originals. QLoRA's paper uses $r{=}64$ with scaling factor $16$
(a scaling-to-rank ratio of $0.25$ against our $2.0$), bf16 compute, a constant learning rate, and a paged 8-bit optimizer. The
optimizer only affects memory, so we use a fused AdamW for every method. Our
\emph{DoRA} rows sit on the quantized base, which is the accepted \textbf{QDoRA}
form \citep{liu2024dora} and is not comparable to NVlabs' fp16 numbers. LoftQ
uses PEFT's one-shot initialization; its headline gains are at 2--3 bit, so
parity with QLoRA at NF4 is the expected result.

\section{Adapter placement: figure and full table}\label{app:placefig}
\begin{table}[tbp]
\centering\small
\caption{Which layers to adapt, phi-1.5 (8-task avg, \%).}
\label{tab:e3}
\begin{tabular}{llccc}
\toprule
Placement & budget $\alpha$ & mean block & seed 42, $+$rs & seed 43, no rs \\
\midrule
near-zero frac (ours) & 0.5 & 12.5 & 73.6 & 72.4 \\
weight-norm & 0.5 & 12.5 & 74.2 & 73.0 \\
depth-matched random & 0.5 & 12.2, 12.6 & 74.3 & 73.3 \\
random & 0.5 & 10.9, 11.3 & 74.6 & 73.5 \\
\midrule
near-zero frac (ours) & 0.7 & 13.4 & 74.8 & 73.7 \\
weight-norm & 0.7 & 13.4 & 75.1 & 73.8 \\
random & 0.7 & 10.8, 11.7 & 75.1 & 74.2 \\
\midrule
depth (top blocks) & 0.7 & 15.2 & 71.2 & n/a \\
depth (top blocks) & 0.85 & 13.3 & 74.6 & n/a \\
near-zero frac (ours) & 1.0 & 11.5 & 75.6 & n/a \\
\bottomrule
\end{tabular}
\\[2pt]
{\scriptsize $\rho{=}.2$. The two columns differ in seed and in rsLoRA scaling, so
compare within a column. ``Mean block'' is the average block index of the adapted
layers, 0 to 23; random draws a new set per seed, so it has two. A second
depth-matched draw without scaling, at seed 42, scores $73.5$. Without scaling,
depth at $\alpha{=}.85$ is the shipped \method{}-s ($73.2$) and $\alpha{=}1.0$ is
\method{}-q ($74.9$ at seed 42).}
\end{table}
\begin{figure}[tbp]
  \centering
  \includegraphics[width=0.55\textwidth]{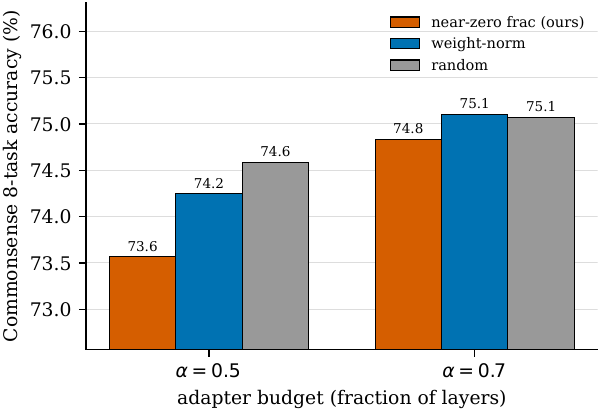}
  \caption{Layer-selection rules against random, two budgets (phi-1.5, seed 42 with
rsLoRA scaling;
Table~\ref{tab:e3} has both seeds).}
  \label{fig:placement}
\end{figure}
\section{Lever ladder}\label{app:leverfig}
\begin{table}[t]
\centering\small
\caption{Speed levers on phi-1.5 (220 steps, V100). GC is gradient
checkpointing.}
\label{tab:ladder}
\begin{tabular}{llcccc}
\toprule
& Configuration & steps/s & tok/s & pad waste & peak (GiB) \\
\midrule
\multirow{6}{*}{\rotatebox{90}{\scriptsize universal}}
& QLoRA, eager attention            & 0.228 & 1{,}046 & 39.8\% & 3.79 \\
& \ + SDPA                          & 0.231 & 1{,}060 & 39.8\% & 3.79 \\
& \ + dropout 0                     & 0.216$^{\dagger}$ & 993 & 39.8\% & 3.79 \\
& \ + length-grouped batching       & 0.245 & 1{,}124 & \best{3.7\%} & 3.79 \\
& \ + fused AdamW (= QLoRA best)    & 0.247 & 1{,}134 & 3.7\% & 3.83 \\
& \ + no GC                         & 0.399 & 1{,}833 & 3.7\% & 18.45 \\
\midrule
\multirow{5}{*}{\rotatebox{90}{\scriptsize \method{}}}
& sparsity $\alpha{=}1.0$ ($+$rs, levers) & 0.260 & 1{,}196 & 3.7\% & 4.05 \\
& depth $\alpha{=}.7$, no truncation      & 0.287 & 1{,}319 & 3.7\% & 4.02 \\
& depth $\alpha{=}.7$ $+$ truncation      & \best{0.303} & \best{1{,}392} & 3.7\% & 4.02 \\
& depth $\alpha{=}.5$ $+$ truncation      & 0.338 & 1{,}554 & 3.7\% & 4.00 \\
& depth $\alpha{=}.7$ $+$ trunc., no GC   & \best{0.568} & \best{2{,}609} & 3.7\% & 14.10 \\
\midrule
& fp16 LoRA (levers)                & 0.339 & 1{,}558 & 3.7\% & 5.18 \\
& rsLoRA (levers)                   & 0.242 & 1{,}112 & 3.7\% & 3.83 \\
\bottomrule
\end{tabular}\\[2pt]
{\scriptsize These are a single sequential pass, used to attribute the speedup to
its parts; the measured comparisons are Tables~\ref{tab:cleantime}
and~\ref{tab:sessions}. The ladder uses depth $\alpha{=}.7$ for attribution; the
shipped speed setting is $\alpha{=}.85$. The \method{} rows include rsLoRA scaling, a scalar on the
update that does not change step time. $^{\dagger}$This row came out slower than
the one above, within single-pass run-to-run noise.
}
\end{table}
\begin{figure}[t]
  \centering
  \includegraphics[width=0.66\textwidth]{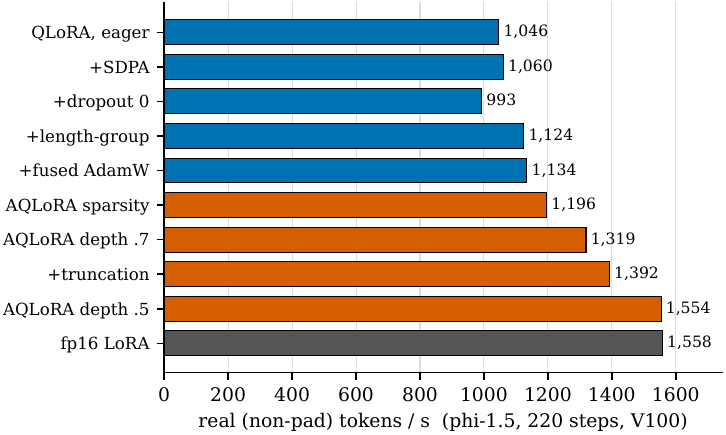}
  \caption{Throughput on phi-1.5 as levers are added. Blue: shared levers;
  orange: \method{}'s own parts; grey: fp16 LoRA.}
  \label{fig:levers}
\end{figure}
\section{Full lever data}
\label{app:levers}
Table~\ref{tab:lever-full} lists all sixteen configurations (phi-1.5, 220 steps, one V100, one sequential pass): the fourteen lever rungs and two
edge profiles that are the source for Table~\ref{tab:ladder}.

\begin{table}[t]
\centering\small
\caption{Full lever data on phi-1.5 (measured): fourteen lever rungs and two
edge profiles.}
\label{tab:lever-full}
\begin{tabular}{lcccc}
\toprule
Configuration & steps/s & real tok/s & pad waste & peak GiB \\
\midrule
QLoRA, eager attention & 0.228 & 1,046 & 39.8\% & 3.79 \\
\quad + SDPA & 0.231 & 1,060 & 39.8\% & 3.79 \\
\quad + dropout 0 & 0.216 & 993 & 39.8\% & 3.79 \\
\quad + length-grouped batching & 0.245 & 1,124 & 3.7\% & 3.79 \\
\quad + fused AdamW (= QLoRA best) & 0.247 & 1,134 & 3.7\% & 3.83 \\
\quad + no grad-ckpt & 0.399 & 1,833 & 3.7\% & 18.45 \\
AQLoRA all-layer 1.0 (paged 8-bit opt) & 0.238 & 1,093 & 39.8\% & 4.01 \\
AQLoRA all-layer 1.0 + levers & 0.260 & 1,196 & 3.7\% & 4.05 \\
AQLoRA depth .7, no truncation & 0.287 & 1,319 & 3.7\% & 4.02 \\
AQLoRA depth .7 + truncation & 0.303 & 1,392 & 3.7\% & 4.02 \\
AQLoRA depth .5 + truncation & 0.338 & 1,554 & 3.7\% & 4.00 \\
AQLoRA depth .7 + trunc, no GC & 0.568 & 2,609 & 3.7\% & 14.10 \\
fp16 LoRA + levers & 0.339 & 1,558 & 3.7\% & 5.18 \\
rsLoRA + levers & 0.242 & 1,112 & 3.7\% & 3.83 \\
\midrule
AQLoRA edge (bs1/seq256/r4) & 0.154 & 232 & 0.0\% & 1.44 \\
QLoRA edge (bs1/seq256/r4) & 0.124 & 187 & 0.0\% & 1.24 \\
\bottomrule
\end{tabular}

\end{table}

\section{Protection budget sweep}\label{app:budget}
\begin{table}[tbp]
\centering\small
\caption{Protection budget swept end to end, phi-1.5.}
\label{tab:rho-sweep}
\begin{tabular}{lccccc}
\toprule
$\rho$ & fp16 layers & base\gib & peak\gib & 8-task avg & step time \\
\midrule
$0$ (QLoRA) & 0  & 0.95 & \best{3.83} & 74.75 $\pm$ 0.05 & ref \\
$.05$       & 7  & 1.01 & 3.89 & 74.97 & $+4.1\%$ \\
$.10$       & 14 & 1.05 & 3.93 & \best{75.08} & $+4.8\%$ \\
$.15$       & 22 & 1.14 & 4.01 & 74.48 & \\
$.20$       & 29 & 1.18 & 4.05 & 74.78 $\pm$ 0.39 & \best{$+5.2\%$} \\
\bottomrule
\end{tabular}\\[2pt]
{\scriptsize Rows $\rho{=}0$ and $\rho{=}.20$ are means of seeds 42 to 44; the
other rows are seed 42. Step time is from one session, relative to QLoRA, with a
$2.8\%$ floor.}
\end{table}

Table~\ref{tab:rho-sweep} sweeps the protection budget on phi-1.5. The three
budgets we timed, in one session,
span $1.1\%$, well inside that session's $2.8\%$ floor, so we cannot separate them on speed and do
not claim a trend.

This result does not carry to every setting. The sweep above is phi-1.5 with the
quality setting. Under the rsLoRA composition the budgets stop being interchangeable, in a way one
seed each cannot resolve. On Phi-3-mini the cheap budget is also worse:
$\rho{=}.05$ scores $86.78$ against $87.04$ at $\rho{=}.20$ and $86.99$ for
QLoRA, over three seeds with standard deviations near $0.15$. The drop is small and inside its own spread, but it is in the wrong direction,
so $\rho{=}.05$ is not a default. The narrower statement holds: the budget is worth sweeping per model, and on the model swept in full the cheap
end lost nothing. The main tables report $\rho{=}.20$ because
that is the budget the rest of the experiments used.

Appendix~\ref{app:rslora} covers the rsLoRA composition.

\section{Error ranking against matched random, per setting}\label{app:nullfull}
Every row holds the protected parameter total equal to top-$K$'s to within
$0.1\%$, so the memory is matched and only the selection differs. Perplexity is
base WikiText-2 with no adapters, which is deterministic given the layer set.
The count $K$ is the size of the top-$K$ set; the matched draws vary in count
because they are matched on parameters instead.

\begin{table}[tbp]
\centering\small
\caption{Error ranking against parameter-matched random draws, per setting.}
\label{tab:ranking-null-full}
\begin{tabular}{lcccccc}
\toprule
Model & $\rho$ & $K$ & top-$K$ & bottom-$K$ & matched random (min/med/max) & draws $<$ top \\
\midrule
phi-1.5 & $0.10$ & 14 & $18.969$ & $18.974$ & $18.911$ / $18.936$ / $18.942$ & 10/10 \\
phi-1.5 & $0.20$ & 29 & $18.935$ & $18.918$ & $18.871$ / $18.906$ / $18.941$ & 8/10 \\
phi-1.5 & $0.30$ & 43 & $18.864$ & $18.869$ & $18.833$ / $18.884$ / $18.909$ & 3/10 \\
Llama-3.2-1B & $0.10$ & 11 & $9.492$ & $9.476$ & $9.463$ / $9.490$ / $9.507$ & 6/10 \\
Llama-3.2-1B & $0.20$ & 22 & $9.457$ & $9.389$ & $9.432$ / $9.471$ / $9.502$ & 3/10 \\
Llama-3.2-1B & $0.30$ & 34 & $9.412$ & $9.268$ & $9.388$ / $9.439$ / $9.457$ & 2/10 \\
Llama-3.2-3B & $0.10$ & 20 & $7.320$ & $7.289$ & $7.317$ / $7.324$ / $7.335$ & 1/10 \\
Llama-3.2-3B & $0.20$ & 39 & $7.308$ & $7.249$ & $7.241$ / $7.310$ / $7.320$ & 4/10 \\
Llama-3.2-3B & $0.30$ & 59 & $7.289$ & $7.208$ & $7.200$ / $7.279$ / $7.296$ & 8/10 \\
Phi-3-mini & $0.10$ & 13 & $5.647$ & $5.677$ & $5.605$ / $5.657$ / $5.665$ & 3/10 \\
Phi-3-mini & $0.20$ & 26 & $5.602$ & $5.658$ & $5.607$ / $5.629$ / $5.638$ & 0/10 \\
Phi-3-mini & $0.30$ & 38 & $5.538$ & $5.626$ & $5.544$ / $5.592$ / $5.625$ & 0/10 \\
Qwen2.5-7B & $0.10$ & 20 & $6.283$ & $6.287$ & $6.288$ / $6.292$ / $6.294$ & 0/10 \\
Qwen2.5-7B & $0.20$ & 39 & $6.271$ & $6.277$ & $6.264$ / $6.284$ / $6.290$ & 2/10 \\
Qwen2.5-7B & $0.30$ & 59 & $6.238$ & $6.265$ & $6.240$ / $6.263$ / $6.273$ & 0/10 \\
\bottomrule
\end{tabular}

\end{table}

\section{Timing estimator and exclusion sensitivity}\label{app:sens}

The headline speed numbers use one estimator (mean of the fastest half of the
repeats) and one inclusion rule (duplicate-arm floor under $3\%$).
Table~\ref{tab:sens} recomputes the pooled result under three other estimators,
with the five excluded sessions forced back in, and on two subsets of the kept
sessions: the nine with a duplicated QLoRA arm, and the twelve that waited for
an empty timing GPU before each run. The point estimate moves by at most $1.3$ points and never changes sign or
ordering. What the exclusion rule buys is not the effect but the error bar:
including sessions that fail their own floor more than triples the spread, from
$\pm 2.3$ to $\pm 7.8$ on \method{}-s, without moving the mean beyond its deviation. Every number
recomputes from the released \texttt{timing.json} files.

\begin{table}[tbp]
\centering\small
\caption{Pooled speed vs.\ QLoRA under alternative timing choices.}
\label{tab:sens}
\begin{tabular}{llccc}
\toprule
Estimator & Sessions & $n$ & \method{}-q & \method{}-s \\
\midrule
fastest half & clean (paper) & 13 & $+4.8 \pm 2.0$ & $+11.3 \pm 2.3$ \\
fastest half & all           & 18 & $+4.5 \pm 4.2$ & $+10.0 \pm 7.8$ \\
plain mean   & clean         & 13 & $+5.0 \pm 2.1$ & $+11.2 \pm 3.4$ \\
plain mean   & all           & 18 & $+5.2 \pm 4.6$ & $+10.3 \pm 8.0$ \\
median       & clean         & 13 & $+5.3 \pm 2.3$ & $+11.6 \pm 4.4$ \\
median       & all           & 18 & $+4.9 \pm 4.4$ & $+10.3 \pm 8.6$ \\
single fastest & clean       & 13 & $+4.2 \pm 2.7$ & $+10.8 \pm 2.1$ \\
single fastest & all         & 18 & $+4.7 \pm 4.2$ & $+10.6 \pm 4.4$ \\
\midrule
fastest half & duplicate arm only & 9 & $+5.4 \pm 1.9$ & $+11.7 \pm 2.3$ \\
fastest half & empty-GPU check only & 12 & $+4.7 \pm 2.0$ & $+11.1 \pm 2.4$ \\
\bottomrule
\end{tabular}\\[2pt]
{\scriptsize ``All'' adds the five floor-failing sessions, including the starved
two-repeat Phi-3-mini session. The last two rows are subsets of the thirteen
kept sessions.}
\end{table}

\end{document}